# USING NLU IN CONTEXT FOR QUESTION ANSWERING: IMPROVING ON FACEBOOK'S bAbI TASKS

**John Ball**

Founder, Pat Inc.
2345 Yale St, 1F
Palo Alto, CA 94306, USA
{john}@pat.ai

## ABSTRACT

For the next step in human to machine interaction, Artificial Intelligence (AI) should interact predominantly using natural language because, if it worked, it would be the fastest way to communicate. Facebook's toy tasks (bAbI) provide a useful benchmark to compare implementations for conversational AI. While the published experiments so far have been based on exploiting the distributional hypothesis with machine learning, our model exploits natural language understanding (NLU) with the decomposition of language based on Role and Reference Grammar (RRG) and the brain-based Patom theory. Our combinatorial system for conversational AI based on linguistics has many advantages—passing bAbI task tests without parsing or statistics while increasing scalability. Our model validates both the *training* and *test* data to find 'garbage' input and output (GIGO). It is not rules-based, nor does it use parts of speech, but instead relies on meaning. While Deep Learning is difficult to debug and fix, every step in our model can be understood and changed like any non-statistical computer program. Deep Learning's lack of explicable reasoning has raised opposition to AI, partly due to fear of the unknown.
To support the goals of AI, we propose extended tasks to use human-level statements with tense, aspect and voice, and embedded clauses with junctures; and answers to be natural language generation (NLG) instead of keywords.
While machine learning permits invalid training data to produce incorrect test responses, our system cannot because the context tracking would need to be intentionally broken. We believe no existing learning systems can currently solve these extended natural language tests. There appears to be a knowledge gap between today's NLP researchers and linguists, but ongoing competitive results such as these promise to narrow that gap.

## 1 INTRODUCTION

The Facebook AI Research (FAIR) bAbI (Weston et al., 2015) is currently a system of 20 tasks that respond to written natural language input to demonstrate understanding in conversation. While most NLP systems actively developed today are based on corpus linguistics, distributional semantics, or combinations of the two, we believe that inadequacies are unresolvable for truly effective Natural Language Understanding (NLU). NLU requires that the meaning of words in sentences are resolved based on the meanings of the other words in the sentence. Each and every word should be understood unambiguously to the extent necessary to converse, despite the ambiguity of the words and phrases in a language.

This paper reports on our team's results in the bAbI tasks undertaken so far to benchmark our NLU system in which context is represented with a list of layered, language-independent semantic sets. The full results are available in spreadsheet form showing the system responses to each of the 10,000 bAbI task training samples[1].

The brain-based Patom theory (Ball, 2017) differs from typical AI systems because it is relatively constraining. Rather than allowing general algorithms to direct the analysis, it allows only sets and lists of patterns as its data structure, and those patterns being stored, matched and used only—in a hierarchy. While a modern digital computer is ill suited for parallel processing, the Patom model is well suited to parallel **matching** using simple hardware (although with an implementation of 150,000 meanings in the system, the possible phrase matches at any point are typically very low, only from 0 to 10).

Sequences and sets are an obvious part of language—with sequential phonemes making up spoken language, and sequential symbols making up written language. These sequences of elements are patterns well suited for matching if their constituents can be matched and, when matched, their use is to access known words. Patom theory uses an identical method to match words from letter sequences as to match phrases from word sequences. The design of the sequence matcher indexes learned patterns of sets of sequential elements through an association with the matched element; i.e., words are matched from their letters with letter sequences as associated stored patterns, and identically, phrases are matched from their patterns stored in constituent words.

Again, these are not just symbols in the patterns, but sets to allow subset recognition. This allows noisy data to still match patterns as there is no fixed pattern starting point and, as the matches are known in advance through learning by experience, missing elements can still be recognized with high accuracy (as the full pattern is already stored).

While today's application of neural networks embraces a number of ancient linguistic models, the modern linguistic framework, Role and Reference Grammar (RRG) (Van Valin, 2005), not only challenges core features of syntactic theory, but also integrates its model with semantic representations and discourse pragmatics. RRG is a highly beneficial model to apply to conversational AI as a result, especially as it is still under active development globally in the linguistic community.

## 2 MODEL

There is the potential for a revolution in language-based AI systems that use our approach because it provides a solution to a number of otherwise open scientific problems. Both AI and cognitive science were founded in 1956; AI at Dartmouth with Minsky, McCarthy and others in the summer; and cognitive science at MIT including Miller and Chomsky in September (Gardner, 1985, 28-30). Both approaches following from there have seemingly exhausted the statistical, artificial neural-network and rules-based approaches, without approaching the requirements of NLU-based systems. Meaning has not been exploited, yet.

Appendix A Science and Engineering Summary, summarises our use of the brain-based Patom Theory with the linguistic framework, Role and Reference Grammar (RRG).

The method uses three software components[2]:
a) The matcher. This takes the input, converts it to known words, and connects to the associated stored meanings (word senses). Phrases connected to the words and senses are tested for matches until no more matches are found. The matcher returns a set of valid, matched predicates (semantic representations/logical structures).
b) The context tracker extracts the logical structures matching the input (and other attributes) and either:
   i) adds to the context sequence if a <u>statement</u> or <u>command</u> was received, or
   ii) finds all valid stored context items if a <u>question</u> was received, with the question type

---

[1] Refer to https://pat.ai/babiresults for the exported results.
[2] Similar to the bAbI reported I.G.O.R. model, except ours uses the semantic representation from the linker's output at the 'I' stage (Weston et al., 2015, 12).

(polar/content) determining the answer[3]

c) For answering, the matched context items are passed to the target language generator for response. The language generator finds a matching phrase and attribute combination and passes the meaning packet including the logical structure, which in turn, based on the requirements of the target language, converts the meaning into text.

Our generic context tracking engine was used unchanged, once a bAbI task's predicates and referents were defined. The generic function is to simply (a) store statements and commands as context and (b) respond to a question using any matches with context——the matches provide the content for natural language generation (NLG). That method is valid for a large class of question-answering cases.

In general, only minor changes to context tracking were required to deal with the different real-world handling of a task, and then additional changes to deal with the required test answers that deviate from human-like responses.

To deal with the real world, little change was required after understanding the responses necessary. One case is list-like. Every item in the list of matching context is returned. Another case is set-like. The last element ONLY is returned (a set of meaning). The present tense, for example, returns the set version "Beth went to the kitchen. Then she went to the garden. Where is Beth?" A: "in the garden". But with past tense, "Where was Beth?", the full list is the response. A: "in the garden and in the kitchen".

## 3 RELATED WORK

Patom theory has never before been used in commercial AI applications because it is new. RRG is today an element in a number of systems around the world. Parsers have been built to exploit the beneficial computational features of RRG (Guest, 2009), but current industry work is on statistical and neural-network models based on Chomsky's syntactic argument[4]. While the solution described by this paper doesn't have a parse step (it is a system that links words to meanings before combining known patterns to resolve predicates), a scaling discussion is included in the appendices.

Other commercial systems such as ARTEMIS and FunGramKB (Cortés-Rodríguez, 2016) have integrated aspects of RRG. The full exploitation of RRG has yet to occur because the research focus has been on other approaches.

## 4 EXPERIMENTS

Nine bAbI tasks were attempted (of 20 available) and, with the debatable exceptions noted below, all passed with 100% accuracy[5]. The design for each test is explained below. The format of the answers produced depend on whether the question is polar or content.

POLAR ANSWERS

To answer a polar question, the question itself is first treated as a statement (The English illocutionary force operator—question/statement/imperative—is just a pattern we can ignore for this purpose). After matching, a yes/no answer is qualified with one or more answers being 'yes'. Rather than just answering yes/no, or affirmative/negative as in bad sci-fi movies, the psa (per RRG) and the auxiliary (aligned for voice) may also be returned, or the full statement, as shown below.

For example: INPUT —Beth went to the kitchen. Is Beth in the kitchen?
Answer:
    a. Yes.

---

[3] We have also validated our technique in answering multiple question cases, like: 'who gave what'.
[4] The argument that syntactic structures should be studied independently from meaning (covered in appendices)
[5] It is qualified what a 'pass' is. While bAbI specifies its answers, they are often not what humans would answer. For example, in one test, bAbI requires one answer (the latest) when two or more are really 'correct'.

    b. Yes, she is.      or      She is.      or      She is, yes.
    c. Yes, she is in the kitchen. or She is in the kitchen. or She is in the kitchen, yes.

Note that discourse pragmatics allows the narrow focus answer to be unambiguously responded with pronouns, even if the answer were "Yes, it broke it" because the actor/undergoer is known unambiguously by both parties based on the question.

### CONTENT ANSWERS

The answers to a content question are different, as the narrow focus question word (who, what, when, where, how, why or referent query which, how many…) logically extends the question. As a result, the replacement for the question word is sufficient, but extension is allowed for clarity.

For example: INPUT—Beth went to the kitchen. Who is in the kitchen?
Answer:
    a. Beth.      or      It is Beth.
    b. Beth is.
    c. Beth is in the kitchen.      or      It is Beth in the kitchen.

If the original referent is qualified as an embedded proposition, at the moment the answer will generate the equivalent qualifications in the answer.

For example: INPUT—The woman who went to the kitchen went to the garden. Who is in the garden?
Answer:
    a. The woman who went to the kitchen, or
        It is the woman who went to the kitchen.
    b. The woman who went to the kitchen is.
    c. The woman who went to the kitchen is in the garden, or
        It is the woman who went to the kitchen in the garden.

These are not stored text answers, but the result of passing the logical structure of an embedded proposition to the language generator for NLG to populate with words.

### 1.1. Task Scale

We extracted the details of each task to ensure all words were included in our vocabulary as well as all predicates were correctly defined. Table 1 (below) shows the metrics for task 5's training set.

     2,000 conversations in the source text of 64,532 lines.
     Vocabulary used:

### Table 1. Vocabulary Extraction bAbI Task 5 Training Set

| 1. Word: Bill | 2. Word: travelled | 3. Word: to |
|---|---|---|
| 4. Word: the | 5. Word: office | 6. Word: picked |
| 7. Word: up | 8. Word: football | 9. Word: there |
| 10. Word: went | 11. Word: bedroom | 12. Word: gave |
| 13. Word: Fred | 14. Word: What | 15. Word: did |
| 16. Word: give | 17. Word: handed | 18. Word: Jeff |
| 19. Word: back | 20. Word: Who | 21. Word: received |
| 22. Word: got | 23. Word: milk | 24. Word: garden |
| 25. Word: hallway | 26. Word: journeyed | 27. Word: moved |
| 28. Word: bathroom | 29. Word: Mary | 30. Word: kitchen |

| 31. Word: took | 32. Word: apple | 33. Word: left |
| 34. Word: passed | 35. Word: put | 36. Word: down |
| 37. Word: grabbed | 38. Word: dropped | 39. Word: discarded |

1.2.     Task Results

The summary results for the tests are shown in Table 1 below. Nine tasks have been attempted to date, with all receiving 100% correct results, except as discussed (where the validity of the target results are challenged).

The Facebook AI Research (FAIR) team are currently looking to improve the machine learning performance for their recurrent entity network (EntNet) solution so it is more effective against the original target of 1,000 cases[6], maximum, instead of 10,000 as used[7]. In contrast, there is no learning step in the method we use other than the definitions of patterns in our semantic network.

### Table 2. Summary Results

| No | Name | Our Results | Testing as of July, 2017 compared to FAIR Feb 2015 Report | | | March 2017 FAIR & Courant Institute |
|---|---|---|---|---|---|---|
| | | | Language Content Loading Required | MemNN — Adaptive Mem + N-grams + Nonlinear Results | Training Required | Recurrent Entity Net Error Rates on 10k Samples of Training |
| 1 | Single Supporting Fact | 100% | 19 words + associations | 100% | 250 examples | 0% |
| 2 | Two Supporting Facts | | | 100% | 500 examples | 0.10% |
| 3 | Three Supporting Facts | | | 100% | 500 examples | 4.10% |
| 4 | Two Argument Relations | | | 100% | 500 examples | 0% |
| 5 | Three Argument Relations | 99.5% | 10 words + associations | 98% | 1000 examples | 0.30% |
| 6 | Yes/No Questions | 100% | 14 words + associations | 100% | 500 examples | 0.20% |
| 7 | Counting | 100% | 4 words + associations | 85% | 10,000 examples | 0% |
| 8 | Lists/Sets | 100% | 1 word + associations | 91% | 5,000 examples | 0.50% |
| 9 | Simple Negation | 100% | 4 words + associations | 100% | 500 examples | 0.10% |
| 10 | Indefinite Knowledge | | | 98% | 1,000 examples | 0.60% |
| 11 | Basic Coreference | 100% | 2 words + associations | 100% | 250 examples | 0.30% |
| 12 | Conjunction | 100% | Zero | 100% | 250 examples | 0% |

---

[6] The target was noted to be set at 1,000 training examples and 95% accuracy arbitrarily (Weston et al, 2015, 6), anyway, but that only applied to systems that use machine learning of that nature, of course.

[7] The comparison between the latest algorithm needing 10k training cases instead of the 1,000 arbitrarily set is included in the the ICLR 2017 report (Henaff et al, 2017, 9).

| | | | | | |
|---|---|---|---|---|---|
| 13 | Compound Coreference | 100% | 6 words + associations | 100% | 250 examples | 1.30% |
| 14 | Time Reasoning | | | 99% | 500 examples | 0% |
| 15 | Basic Deduction | | | 100% | 100 examples | 0% |
| 16 | Basic Induction | | | 100% | 100 examples | 0.20% |
| 17 | Positional Reasoning | | | 65% | >10,000 examples | 0.50% |
| 18 | Size Reasoning | | | 95% | 1,000 examples | 0.30% |
| 19 | Path Finding | | | 36% | >10,000 examples | 2.30% |
| 20 | Agent's Motivations | | | 100% | 250 examples | 0% |

1.3. Task 1 Single Supporting Fact

1.3.1. Results
100%

1.3.2. Sample

> **Task 1: Single Supporting Fact**
> Mary went to the bathroom.
> John moved to the hallway.
> Mary travelled to the office.
> Where is Mary? A: office

1.3.3. Discussion

This task tests the recall of a mentioned destination, converted to a position. But by asking for a one-word answer, the meaning of the answer is replaced with a keyword. Linguistically, the answer 'bathroom' should be substitutable into the question after changing the illocutionary force operator from question to statement. ("Mary went to the bathroom. Where is Mary?" -> *Mary is bathroom. But bathroom is not a position like "here", but the object of the preposition whose predicate represents "where".) An AI test needs to deal in real language, not keywords, to allow natural interaction.

Our NLU first converts "Where is Mary?" to "Mary is where?" in context. Next, because it is a question, we intersect our context to find positions of Mary (solve for where, a meaning category). Then we pass the latest position to the NLG, which generates a position; e.g. "in the kitchen" since that is the representation of the position (a two-place predicate, 'to' with one place necessary to answer "where"—the first argument). 'to' is defined with a number of ambiguous representations that with an argument of "kitchen", a 3D location, tells the generator to use 'in'. (The possible meanings of 'to' is disambiguated in the NLU step when the predicate "in the kitchen" is matched.)

These predicates (went, travelled, moved) extend a motion (go) into an active achievement. By comparison, 'is position' is a state, not ingressive (INGR), so it positions something in space. The destination applies to motion only (e.g. *I slept to the kitchen), so the predicate type must be validated in the proposition pattern during the matching step. Note: As contrast, "I slept in the kitchen" positions the act of sleeping in space.

The two logical structures making up the context, an active achievement[8], are ignoring the method of motion for now because it doesn't affect this specific test.
Mary went to the kitchen
do' (mary, [go.motion'(mary)]) & INGR be-in' (the kitchen, mary)

Here, there are two predicates. The first predicate represents the motion undertaken by the actor, Mary ('go' does not encode manner, but the other predicates do). The resulting predicate has the resulting state positioned, Mary being "in the kitchen". Note that the object of the preposition constrains the meaning of the preposition as with any predicate. (Predicates constrain the meaning of their arguments when matched, making the language less ambiguous.) Going to the mat means you are ON the mat at the end of the motion. Going to the beach means you are AT the beach or ON the beach.

The state version (motion is an activity verb class) has the following example. (Note the position of the object takes the first argument slot.)
Mary is in the kitchen
be-in' (the kitchen, mary)

The rest of the context received includes the operators and attributes matched in the source sentence—like tense, aspect, negation and grammatical elements like voice.

The context handing is quite simple. First, intersect[9] the context with the question's logical structure:
Where is Mary
be-LOC' (Where, Mary)

Note that although 'where' is a predicate, it is shown as two elements, but a 'where' statement can be responded to with a prepositional phrase or deictic (e.g. here, there). There is probably a better way to show this logical structure. We're on it.

Second, to answer the question, two intersected matches are returned, and the correct answer is <u>the</u> <u>last</u> <u>one</u> in context (as present tense is a state, and so the answer is the current state). Why? Because positions take place over time, and we are only at the latest one in the real world. This process is simplified due to artificial restrictions on the test for tense. Given the question LS and the answer LS, 'where' generates to "in the kitchen" given the match and ignoring the match with 'Mary' shown as ∅ to leave focus on the contributing parts:
be-in' (the kitchen, ∅)

1.3.4. Suggested Improvements

The easiest way to improve this test is to replace the keyword response with an English language response. So instead of the word 'kitchen', the answer should be "in the kitchen". Examples showing the recognition of the location would mix the necessary preposition with the valid object (based on its spatial, dimensional characteristics).

Another improvement is to embed meanings: "**The woman who went to the garden** went to the kitchen. Where is the woman? **IN** the kitchen."

The addition of tense, aspect, voice and pronoun use would make the tests more natural. The woman WAS going to the kitchen. She is in the garden now. Where is she? Where was she? Note: "Where **is** the woman?" returns a single position: "in the kitchen". "Where **was** the woman?" returns a list which is responded to in English as a sequence separated with a final conjunction (Where was the woman? She was "in the kitchen and in the garden").

---

[8] Technically, the literature calls this an active accomplishment (Van Valin, 2005, 44-45). I prefer the terminology that aligns with INGR—an achievement (Pavey, 2010, 100) to keep things simple for students.

[9] Patom theory describes *linkset intersection* as the combination of sets with a union, leaving only sets that match the constituents of the elements. A meaning-based system has a number of different types of interactions based on the semantic relations.

Again, tense is treated poorly in bAbI now, usually adding propositions in the past tense and asking questions in the present or present progressive. Normal speakers might answer differently, for example: "Mary went to London. Where is Mary?" A: "I don't know, but she **was** in London." Our system responds with the source tense for this reason in our spreadsheet of results.

---

1.4. Task 6 Yes/No questions
  1.4.1. Results
      100%
  1.4.2. Sample

> **Task 6: Yes/No Questions**
> John moved to the playground.
> Daniel went to the bathroom.
> John went back to the hallway.
> Is John in the playground? A: no
> Is Daniel in the bathroom? A: yes

  1.4.3. Discussion
      In Hollywood, polar questions (yes/no) have traditionally been answered by robots and logical aliens with "affirmative" or "negative", but humans say things like, "Yes, he did" and "No, he didn't do it". Today with AI, the longer a phrase is, the less likely the machine understands. With people, the opposite is true: more words means more accuracy because repetition—by saying yes and an equivalent answer—is better since even if you miss a word, you still receive the response. We also extend a negative response with a clarifying point, if we know it, as in: "No, but Susan ate it".

      Of interest with relation to tenses, for the example: "Daniel went to the bathroom. Where is Daniel?", a human might say, "I don't know" or "I'm not sure." Tense is a powerful communications tool, and just because he went there doesn't mean he IS there. Our response to such a mixed tense question-answer pair is to clarify with, "Yes, he WAS there."

      Note that the logical structures for these questions are the same as in Task 1, and so examples are not repeated here.

  1.4.4. Suggested Improvements
      Convert the answers to include clarifications and contrasts as described in the discussion points. To quickly validate the answer, we set the testing randomizer to respond with a specific template starting with the word 'yes' or 'no' and then a random selection of clarifications. Dependency injection tools make this simple in modern programming environments without affecting production responses.

      A use of tense (and pronouns) also makes for more natural dialog; for example, "John went to the playground. Then he went to the office. He is in the garden now. Where is John? (in the garden). Where was John? (in the playground and in the office)." This can either leave out the current position, or include it.

      The use of embedded phrases also makes the dialog more natural; e.g., "John who went to the playground went to the office. He is in the garden now." This produces the same results because an embedded element of context is entered into tracking first,

as it is a constituent of the whole proposition.

---

1.5. Task 9 Simple Negation
  1.5.1. Results
    100%
  1.5.2. Sample

> **Task 9: Simple Negation**
> Sandra travelled to the office.
> Fred is no longer in the office.
> Is Fred in the office? A: no
> Is Sandra in the office? A: yes

  1.5.3. Discussion

  This task is the same as Task 1, but with the inclusion of negation and the expression 'be no longer'. By adding a be-no-longer case as 'was' AND then 'is NOT', the context is resolved without change.

  First, the positive and negative positions and destinations are added into context. Then, as only present tense questions are asked, we use linkset intersection to produce a shortlist of valid answers, and then respond with the last one.

  As the questions are polar, 'yes' answers include a supporting phrase; e.g., "Yes, she is" in which the auxiliary in the question provides the additional validation since the question provides the narrow-scope template we respond with. In the case of negation, a similar validation can be added or, where a similar match is found that differs from the intersected version, that validation is offered.

  For example: "Sandra travelled to the office. Is Fred in the office? No, but Sandra is."

  Note that the logical structures for these questions are the same as in Tasks 1 and 6, and so examples are not repeated here.

  1.5.4. Suggested Improvements
    This task can be rolled up into a multi-tense task, including past, present and future.

---

1.6. Task 11 Basic Coreference
  1.6.1. Results
    100%
  1.6.2. Sample

> **Task 11: Basic Coreference**
> Daniel was in the kitchen.
> Then he went to the studio.
> Sandra was in the office.
> Where is Daniel? A: studio

  1.6.3. Discussion

  This test leverages a simplistic coreference model in which the last mentioned referent's pronoun can be used for additional context tracking.

  There are many forms of coreference in language—when (soon, later, now), where (here/there), who (he/she/it), whom (him/her/it) and so on. Perhaps these tests can extend the use further by including unambiguous cases—mixing the order to produce a natural flow.

1.6.4. Suggested Improvements

The use of the 'adverbial'/temporal connecting words makes the text far more readable; but the test ignores those elements. The joining of such propositions would make the test more challenging: "Daniel was in the studio before he went to the kitchen." This is just a temporal sequence and equivalent to the sample. Perhaps Task 14 covers this somewhat, but the awkward phrases make for tough reading.

---

1.7. Task 12 Conjunction

1.7.1. Results

100%

1.7.2. Sample

> **Task 12: Conjunction**
> Mary and Jeff went to the kitchen.
> Then Jeff went to the park.
> Where is Mary? A: kitchen
> Where is Jeff? A: park

1.7.3. Discussion

Conjunctions provide a meaning-based system with a question; do you resolve the conjunctions into statements, or leave as a collection of bundled information. We chose to retain the combinations to avoid data corruption. By exploding the permutations, the original message is corrupted. The pronouns 'they'/'them'/'their' rely on the bundle being retained. Also, by creating a state separated from the propositions, clarity can be lost in the translation, so it's better to retain NLU meaning bundles and extract only when questions match.

do' (mary and jeff, [motion'(mary and jeff)]) & INGR be-in' (the kitchen, mary and jeff)

For questions about the mover, there is the pronoun 'they'. For polar questions about Mary's position, the context tracking drills down into the second LS and the undergoer position matches Mary (that is, 'Mary and Jeff' matches 'Mary' through intersection).

1.7.4. Suggested Improvements

The use of conjunctions in the psa (privileged syntactic argument in RRG) can be extended to the destinations (to the kitchen and the office) and other arguments. Languages make extensive use of combinatorial capacity and so such changes must be reflected in a solid AI benchmarking system.

---

1.8. Task 13 Compound Coreference

1.8.1. Results

100%

1.8.2. Sample

> **Task 13: Compound Coreference**
> Daniel and Sandra journeyed to the office.
> Then they went to the garden.
> Sandra and John travelled to the kitchen.
> After that they moved to the hallway.
> Where is Daniel? A: garden

1.8.3. Discussion

Our context tracking system incorporates patterns such as compounds into the referent matching process. So when the referent is matched into the predicate 'journeyed', it does so with third-person plural attributes.

The pronoun for a third-person plural referent in English is 'they' and therefore no additional work was required to deal with this case.

### 1.8.4. Suggested Improvements
Could incorporate into Task 12.

---

Tasks 5, 7 and 8 notes

Possession manipulation: have'(actor, undergoer)

These last three bAbI tasks in our completed set all manipulate possession. The predicate **have'(x,y)** uses a number of alternative forms to test the system:
- polarity (positive/negative); e.g. **NOT have'(x,y)**
- accomplishment; e.g. **BECOME have'(x,y),** and
- causative three-role combinations; e.g. **[do'(actor,∅)] CAUSE [BECOME have'(to,undergoer)]**

There are some additional logical structures that are simple combinations of these elements, such as when you give something to someone, you no longer have it, and then they have it. That is shown with the LS (for "Mary gave the milk to Bill"). NOTE the first element representing the shortened phrase, "Mary gave the milk":

> **[do'(Mary,∅)] CAUSE
> [BECOME NOT have'(Mary,milk) ∧ BECOME have'(Bill,milk)]**

The following table is an initial analysis of the predicates used and their attributes:

### Table 3. Have predicates

| Predicate | No. roles | Type | Extend | Polarity | Cause | Example | LS Cause | LS (base) | LS (extension - to/from) |
|---|---|---|---|---|---|---|---|---|---|
| have, own, possess | 2 | state | - | + | NO | Bill has the milk | | have'(Bill, milk) | |
| pick up | 2 | state, accomplishment | - | + | NO | Bill picked the milk up | | BECOME have'(Bill, milk) | |
| get | 3 | state, accomplishment | from | + | NO | Bill got the milk from Mary | | BECOME have'(Bill, milk) | BECOME NOT have'(Mary, milk) |
| receive | 3 | state, accomplishment | from | + | NO | Bill received the milk from Mary | | BECOME have'(Bill, milk) | BECOME NOT have'(Mary, milk) |
| take | 3 | causative, state, accomplishment | from | + | YES | Bill took the milk from Mary | [do'(Bill,0)] CAUSE [] | BECOME have'(Bill, milk) | BECOME NOT have'(Mary, milk) |
| grab | 3 | causative, state, accomplishment | from | + | YES | Bill grabbed the milk from Mary | [do'(Bill,0)] CAUSE [] | BECOME have'(Bill, milk) | BECOME NOT have'(Mary, milk) |
| leave | 2 | state, accomplishment | - | - | NO | Bill left the milk | | BECOME NOT have'(Bill, milk) | |
| discard | 2 | state, accomplishment | - | - | NO | Bill discarded the milk | | BECOME NOT have'(Bill, milk) | |
| put down | 2 | state, accomplishment | - | - | NO | Bill put the milk down | | BECOME NOT have'(Bill, milk) | |
| drop | 2 | state, accomplishment | - | - | NO | Bill dropped the milk | | BECOME NOT have'(Bill, milk) | |
| gave | 3 | causative, state, accomplishment | to | - | YES | Bill gave the milk to Mary | [do'(Bill,0)] CAUSE [] | BECOME NOT have'(Bill, milk) | BECOME have'(Mary, milk) |
| handed | 3 | causative, state, accomplishment | to | - | YES | Bill handed the milk to Mary | [do'(Bill,0)] CAUSE [] | BECOME NOT have'(Bill, milk) | BECOME have'(Mary, milk) |

Of interest is the reduction in the number of meanings to be tracked by the system. Our initial design was based on the concept of one word, one meaning. bAbI demonstrated that a meaning of 'proximity' plus attributes is sufficient to model destination adpositions in English—a reduction of 12 or so different meanings to one that provides more flexibility.

These possession tasks also were modeled efficiently with a reduction in meanings. Instead of take/drop/give having single meanings, they all consolidated into their base predicate (**have'**) with some additional attributes (as shown in Table 3; e.g. positive/negative, causative).

From a computational perspective, the reduction in meanings results in a reduction in combinations, and therefore a reduction in effort. The networking model (as illustrated with George Miller's WordNet from the 1980s—separating word senses from word forms) allows more specific cases to simply inherit the characteristics of an associated sense such as in the case of the pure motion 'go' providing the specific kind of motion's semantic model, 'journey', which can use the 'entails' association between it and 'go'.

1.9. Task 5 Three argument relations

1.9.1. Results

99.5%

(The 'failing' tests represent (a) an error in the test result due to an error in the input file. We did not attempt to recode to produce the wrong result to pass the test as, without machine learning, it would be complex to code intentional ad hoc errors. There was also (b) an error in interpretation of a logical structure that, after review, we thought worthy of further discussion below.

1.9.2. Sample

> **Task 5: Three Argument Relations**
> Mary gave the cake to Fred.
> Fred gave the cake to Bill.
> Jeff was given the milk by Bill.
> Who gave the cake to Fred? A: Mary
> Who did Fred give the cake to? A: Bill

1.9.3. Discussion

A failed test looks a bit like this: "Mary went to the kitchen. She picked the milk up. Did Mary receive the milk?"
We think she did. Does this have a valid 'no' answer? Why worry? Because she acquired it in the kitchen. Ask your friends what they think. If you say 'no', don't you feel deceptive? Of course there is another sense of 'take' (in which you carry something somewhere) but that's not the case here.

Taking the predicates, the cases of **have'** are (the motion predicates are not questioned, but already set up for other tasks): picked up, gave, handed, received, got, took, left, passed, put down, grabbed, dropped, and discarded.

Our team did this task twice. First, using the meaning **have'**, the forms 'give' and 'take' are negative and positive cases respectively. The tests all passed, until we added the distinctions for Tasks 7 and 8: the achievement condition. So the context tracking was updated to align the causative and accomplishment properties.

Let's look at the results for set 8 to understand a failure.

### Table 4. A 'failed' test case

| ---- SET 8 ---- | | |
|---|---|---|
| Bill grabbed the apple there. Bill got the football there. Jeff journeyed to the bathroom. Bill handed the apple to Jeff. What did Bill give to Jeff? | the apple | Passed |
| Jeff handed the apple to Bill. Bill handed the apple to Jeff. What did Bill give to Jeff? | the apple | Passed |
| Jeff handed the apple to Bill. Bill handed the apple to Jeff. What did Bill give to Jeff? | the apple | Passed |
| Jeff put down the apple. Bill passed the football to Jeff. What did Bill give to Jeff? | the football | Failed |
| Jeff passed the football to Bill. Mary got the milk there. What did Bill give to Jeff? | the football | Failed |

Notice the first error: "What did Bill give to Jeff?" ANSWER: "The football". The expected answer was "the apple". So what went wrong?

Below is the source data for set 8 that includes the answer reference number. Note that line 13 has Bill giving the football to Jeff, but both lines 14 and 17 reference line 10 for its answer, instead of 13.

The better answer, which our system initially had as its default answer, is both: "the football and the apple" corresponding to the statements "Bill handed the apple to Jeff" and "Bill passed the football to Jeff". We think a human answer would also be both.

### Table 5. bAbI Extract Issue

| 1 Bill grabbed the apple there. | | |
|---|---|---|
| 2 Bill got the football there. | | |
| 3 Jeff journeyed to the bathroom. | | |
| 4 Bill handed the apple to Jeff. | | |
| 5 What did Bill give to Jeff? | apple | 4 |
| 6 Jeff handed the apple to Bill. | | |
| 7 Bill handed the apple to Jeff. | | |
| 8 What did Bill give to Jeff? | apple | 7 |
| 9 Jeff handed the apple to Bill. | | |
| 10 Bill handed the apple to Jeff. | | |
| 11 What did Bill give to Jeff? | apple | 10 |
| 12 Jeff put down the apple. | | |
| 13 Bill passed the football to Jeff. | | |
| 14 What did Bill give to Jeff? | apple | 10 |
| 15 Jeff passed the football to Bill. | | |
| 16 Mary got the milk there. | | |
| 17 What did Bill give to Jeff? | apple | 10 |

Looking at line 13, you see the last proposition: "Bill passed the football to Jeff." If Bill passed the football to Jeff, he also gave it to Jeff, with a different manner. A check of the other cases in bAbI shows that analysis to be consistent. Here is set 6, (in table 6 below) which shows on the last line that "Mary passed the football to Fred", which results in the answer: "Who gave the football to Fred?" as "Mary". So 'passing' is definitely 'giving'.

**Table 6. Giving is passing**

| ---- SET 6 ---- | | |
|---|---|---|
| Jeff took the milk there. Jeff gave the milk to Bill. Who did Jeff give the milk to? | Bill | Passed |
| Mary travelled to the bathroom. Mary journeyed to the hallway. Who received the milk? | Bill | Passed |
| Bill went to the kitchen. Fred grabbed the apple there. What did Jeff give to Bill? | the milk | Passed |
| Fred left the apple there. Mary went back to the bathroom. Bill went to the hallway. Bill moved to the kitchen. Bill journeyed to the bathroom. Mary got the football there. Mary travelled to the office. Mary discarded the football. Fred got the football there. Fred gave the football to Mary. Who gave the football to Mary? | Fred | Passed |
| Mary passed the football to Fred. Mary got the apple there. Who gave the football to Fred? | Mary | Passed |

So the bAbI answer in set 8 is simply wrong. But if it is wrong, how can anyone claim 100% passing? AI should be able to check the <u>input</u> as well as the <u>output</u>—an impossibility with the current design of most machine learning approaches.

GIGO happens in these situations. Machine learning is at risk from bad input data.

Since this is our test of the TRAINING data, the training data must be wrong.

Looking for another failure, we see this one:

### Table 7. Another 'failure' discussion

| ---- SET 1446 ---- | EXPECTED | ANSWER | |
|---|---|---|---|
| Bill moved to the office. Fred went back to the bathroom. Mary journeyed to the bathroom. Fred went to the garden. Fred grabbed the milk there. Fred journeyed to the bathroom. Fred put down the milk there. Mary got the milk there. Mary went back to the hallway. Bill travelled to the kitchen. Fred went to the bedroom. Mary dropped the milk. Jeff went back to the bathroom. Mary got the milk there. Jeff journeyed to the kitchen. Mary got the apple there. Fred journeyed to the hallway. Mary gave the apple to Fred. Who did Mary give the apple to? | Fred | Fred | Passed |
| Mary journeyed to the garden. Fred left the apple. What did Mary give to Fred? | apple | the apple | Passed |
| Mary moved to the bedroom. Jeff moved to the bedroom. What did Mary give to Fred? | apple | the apple | Passed |
| Fred took the apple there. Mary gave the milk to Jeff. What did Mary give to Jeff? | milk | the milk | Passed |
| Mary picked up the football there. Mary passed the football to Jeff. What did Mary give to Jeff? | milk | the football | Failed |

Notice on the final line, the expected answer is "the milk", but the NLU output is "the football". Again, the same case is seen: "Mary passed the football to Jeff" means she gave it to Jeff. So it is another error in the training data.

So that's a consistent feature of the training data. We will keep our result as 'Failed' for the moment as the human reaction is that the answer produced is correct if limited to the last case. The more natural case is "the football, the milk and the apple" looking at the source above because 'giving' should return a <u>list</u>, unlike possession of a single object.

There was one more type of error in our tests of the training data. Here is a specific case:

### Table 8. If you take it, did you receive it?

| ---- SET 1477 ---- | EXPECTED | ANSWER | |
|---|---|---|---|
| Bill went to the kitchen. Bill journeyed to the bedroom. Fred grabbed the football there. Fred gave the football to Mary. Who gave the football to Mary? | Fred | Fred | Passed |
| Mary dropped the football. Bill took the football there. What did Fred give to Mary? | football | the football | Passed |
| Fred moved to the kitchen. Bill discarded the football. Who received the football? | Mary | Bill | Failed |
| Jeff grabbed the milk there. Mary grabbed the football there. Mary left the football. Fred journeyed to the bathroom. Jeff put down the milk. Fred went to the office. Mary went back to the bathroom. Fred went back to the garden. Mary journeyed to the kitchen. Bill grabbed the football there. Jeff went to the bedroom. Fred went to the kitchen. Jeff went to the kitchen. Jeff got the milk there. Jeff gave the milk to Mary. Bill put down the football there. Who did Jeff give the milk to? | Mary | Mary | Passed |
| Bill travelled to the bathroom. Mary left the milk. Who gave the milk? | Jeff | Jeff | Passed |

So what is wrong here? We decided to keep this as a talking point, also, as it highlights a feature of the meaning-based model. Reading the bAbI guide again, it could be read that you cannot give something to yourself, nor can you take something from yourself. But "John took the football from himself" and "Beth gave the football to herself" are understandable.

The NLU says that Bill received the football, not Mary as expected by bAbI.

Clearly, the statement "Bill took the football there" means "Bill received the football". The resolution of 'there' is debatable, but not important now. Why does 'take' mean receive? Here's the output from the meaning matcher display:

Input text: Bill took the football there
be-LOC' (there, [do'(bill,∅)] CAUSE [BECOME have' (Bill, the football)])

So based on the logical structure, "Bill took the football" means he gave it to himself.

As an aside, by writing "Bill took the football there" the word sense for 'took' is forced to mean "carried" instead of "received" because of the word "there".

It is a trivial matter to exclude this case, but we think that a human would avoid this by saying it less ambiguously, or making the statement more clearly (Instead of "Bill took the football there" why not "Bill picked the football up there" if the intent is to say nobody caused him to have it). That is a basic rule of linguistics that as the intent of language is to communicate, the transmitter tries to be helpful.

Which brings up the critical point about trusting the decisions of AI, that is such a talking point in the press: "...if you build your system entirely with deep learning… and something goes wrong, it's hard to know what's going on and that makes it hard to debug[10]."

We agree. Fortunately, this meaning-based system is very different. For us, every matched pattern and promoted phrase can be seen and traced back. The path from words to phrases to meaning-based propositions is completely understood, is documented and can be controlled at all levels in our code.

By building a Natural Language Understanding (NLU) machine in which the meaning of the words in a sentence is determined by the meanings of the other words in the sentence, we have total control of the process and total visibility. Semantically, the hypernym/hyponym or 'is-a' relations are well developed. If a doctor is a person, there is no need to decide whether the doctor is male or female. The bias from Deep Learning is a natural consequence of a system that generalizes meaning based on proximity (i.e. cosine distances) rather than what was written. Our system is the opposite of the way almost all the current range of black-box solutions work (for example, word2vec, Deep Learning, Speech to Text) where even a skilled practitioner cannot look at the data and explain how all the proximities were calculated or what the internal details mean.

---

[10] https://techcrunch.com/2017/04/01/discussing-the-limits-of-artificial-intelligence/

- 1.9.4. Suggested Improvements

    The current sample statements take the last correct answer. When someone receives more than one item, only the last case is to be answered to get the correct answer. That is the wrong approach. Let's go with all the cases instead, or clarify with: "Who was last to give the milk."

## 1.10. Task 7 Counting
- 1.10.1. Results

    100%
- 1.10.2. Sample

    **Task 7: Counting**
    Daniel picked up the football.
    Daniel dropped the football.
    Daniel got the milk.
    Daniel took the apple.
    How many objects is Daniel holding? A: two

- 1.10.3. Discussion

    This tasks exploits the polarity of the logical structure (LS) or semantic relationship. Negative polarity reduces by one the number of have' cases. Particle verbs like 'pick up' and 'put down' allow the split between the direct core argument and the particle. "Daniel picked the football up" is an example of a normal, and in some cases, preferred, option to "Daniel picked up the football".

    The question "How many objects is Daniel holding?" begs the follow-on questions like "How many <u>fruits</u> is Daniel holding?" and "How many <u>drinks</u> is Daniel holding?" With linkset intersection, the questioning referent provides 'object' in bAbI, which intersects with just about everything. But with no change to the architecture of a meaning-based system, the other questions would be answered correctly.

- 1.10.4. Suggested Improvements

    The combinations of referent phrases is currently restricted to two (definite determiner plus noun e.g. 'the milk', and proper noun e.g. 'Mary'). We suggest this increases to a simple range like (cardinal plus plural noun e.g. "five footballs" and determiner plus cardinal plus plural noun "the three footballs").[11] Counting becomes less trivial when the test starts with "Daniel picked three footballs up."

    Particle verbs like these can be split—so let's use it. Our system deals with the split by matching that pattern in turn and validating it with its predicate. We are curious how Deep Learning deals with this variation.

    We would also like to see the tasks extended to drill down into the category as mentioned. To allow a training set to determine category will need the addition of classifying statements like: "Milk is a drink. An apple is fruit." and so on, but as the category should be general in conversation, it is worth testing for this type of understanding even though the list may need to be very long to catch the full list of types if the system doesn't rely on an inheritance network (milk is a drink, a liquid, a white fluid…).

## 1.11. Task 8 Lists/Sets
- 1.11.1. Results

---

[11] There are obviously a few cases to deal with depending whether the noun is a count or mass type, but even four or five cases will make the proposition choices much more interesting.

    100%

1.11.2. Sample

> **Task 8: Lists/Sets**
> Daniel picks up the football.
> Daniel drops the newspaper.
> Daniel picks up the milk.
> John took the apple.
> What is Daniel holding? A: milk, football

1.11.3. Discussion

    Task 8 is the same as Task 7, except that the question is for a count of elements on the intersected list for Task 7, instead of the named list of elements for Task 8. Comments for Task 7 therefore apply to this task.

1.11.4. Suggested Improvements

    Per Task 7, but also include the full referent. It is "the football", not just "football". A mass noun like "milk" could also be better qualified, like "some milk" or "a glass of milk". The point is that normally only proper nouns in context stand without a determiner in English in this narrow focus and "football" as an answer, in particular, is not good English.

# 5 DISCUSSION

The goal of the bAbI tasks is to produce the correct answers to questions about a story without 'task-specific engineering' and in so doing, to improve the state of the art to perform more complex solutions within a closed loop (virtuous circle). By increasing the difficulty of the tasks to break the current model, incremental improvement will take place.

Since the initial 20 cases were published, there appears to be no increase in task difficulty, or additional tasks, but our meaning-based method allows context bundles to be embedded in phrases and so a number of extensions have already been tested in our lab.

For example, in our NLU "Mary who went to the kitchen went to the garden. Where is Mary?" returns "in the garden" because the embedded context 'Mary who went to the kitchen' (the referent, Mary) is added first:

    do' (mary, [motion'(mary)]) & INGR be-in' (the kitchen, mary)

before being added to the argument of the destination "to the garden"

    do' (mary, [motion'(mary who went to the kitchen)]) & INGR be-in' (the garden, mary who went to the kitchen).

And of course: "Where <u>was</u> Mary?" A: "in the kitchen".

Even the correct use of position as a phrase should be introduced to bAbI. Given this input: "Mary went to the beach. Mary went to the mat. Mary went to kitchen. Where was Mary". The expected answer to "Where was Mary" would be: "in the kitchen, on the mat and at the beach". It certainly wouldn't be: "kitchen, mat, beach".

While FAIR is using their technology to do new, unrelated tests perhaps more aligned to the strengths of machine learning and Deep Learning, the original goal of bAbI remains a good basis for improved AI.

SIMPLE EXTENSIONS
bAbI tasks appear to be designed around simplification of some of the more productive features of languages. It may be that this was intentional to enable machine learning tasks to compete, initially, but a way to improve the tasks would be for linguists to build the tests without knowledge of the algorithms, or other approach, used to perform the QA. Certainly after its inception in 2015, and another successful report from FAIR on their progress with new algorithms, now is a good time to take the next steps. The following extensions to bAbI radically increase the combinations of possible inputs, but only in line with very simple language. As conversing in language is the goal, a system that copes with more natural language input is necessary. If this causes challenges to the machine learning teams, that is good to re-focus the team on fundamental requirements.

OPERATORS AND OTHER LANGUAGE FEATURES

All bAbI tasks suffer from limited tenses today. Languages use tense, modality, aspect and voice to produce useful communications. Our NLU already deals with the few patterns necessary to deal with these linguistic features. For example: a passive example for a 'have' proposition; instead of "Bill gave the milk to Mary", use "The milk was given to Mary by Bill" or other minor variants like "The milk was given by Bill to Mary" and the unmarked third argument version, "Bill gave Mary the milk" where the word order identifies the argument's roles. Other aspect variations extend the variations further to be more natural, like the 'get' passive, "The milk got given to Mary by Bill" or with progressive, perfect aspect, "The milk had been getting given to Mary by Bill" or the future form of that, "The milk will have been getting given to Mary by Bill".

QUALIA

Qualia uses the meanings of the referents, like 'car', to identify the clear meaning of the sentence. "The car screeched down the road" has the predicate 'screech'. Here, the meaning relates to a part of a car: the wheels. When wheels turn under high acceleration, the rubber tires' interaction with the road creates a screeching noise. Here, the car HAS wheels that HAVE tires (holonym/meronym relations). While language simplifies the communications, a meaning-based system accounts for this as recognized by RRG (Van Valin, 2005, 50-52) and Lexical Semantics (Pustejovsky et al, 1996, 1-13). Even samples like "the car started" are referring to another part of the car—the engine. bAbI should include tests for qualia recognition as it deals with an important conversational element. This will require the associations to be defined in the test, perhaps, to remain in line with the task's spirit, but as a common feature of language, its implementation in NLU products will be appreciated in the future as we take control of our machines, not let them control us.

RE-USE OF METHOD

"Ilikeyou" illustrates another level of AI limitation. The conversion from text (text is made of letters, atomic patterns of meaning encoded in modern computers; e.g., Unicode, ASCII and UTF8) to words and phrases would be expected to follow the same principles in a brain, but in modern AI it does not. The atomic property of letters allows AI researchers to avoid part of the acquisition problem, but gets caught on the resulting problem—the duplication and compression of the letters, as is the compression or exclusion of word meanings. Hashing algorithms can help, but they are not brain-like. The approach of using letters as encoded elements instead of consolidating them as patterns is symptomatic of the difference between computers and brains. While brains seem to use a single method to store multi-sensory patterns and associations, computer programmers in AI start with encoded duplication, and extend from there. In the sentence above, a computer program may need to be written to work out "I"-"like"-"you" (word boundary identification), which should be determined by the fact that those words make sense together, not because those words are in a dictionary or hash table. Other obvious words to exclude from the string are: "il" (French 'he'), "i" (again), "key", "yo". It is the combination of meaning, and the full use of every element of the sequence we have found powerful and effective for our linking algorithm.

Our view is that a single method is better for AI scalability rather than multiple black boxes.

Our work to date has not been widely published as our priority is to create the system and its production applications. At the time of writing, NLU in particular continues to be the focus of many startups around the world, albeit a majority seem to be focussed on statistical approaches like Deep Learning.

## 6  CONCLUSION

bAbI tasks were created to test machine learning systems with a goal of finding ways to improve conversational AI. As a benchmark test, bAbI delivers on that promise further by allowing other forms of AI to compete.

Our model solves a number of problems by: (a) exploiting meaning, using the RRG linking algorithm and set-based simplifications like operators and attributes and (b) removing historically unsuccessful methods, which for decades have been a high priority goal in AI, like parsing and its ambiguous reliance on part-of-speech definitions. The next step is to add complexity to the test, based on linguists blind input, as suggested in this paper, driven by the need for AI to progress against conversational tasks.

Our team appreciates these benchmarks to compel us to decompose our model into more granular elements in line with the virtuous circle and are looking forward to new levels of complexity.

Can the distributional hypothesis deal with the extended tasks and theoretical limitations? Not in their current form as the hypothesis relates to word forms, not meanings, and when the meanings are inferred from other facts not necessarily in the text at all.

There are fundamental limitations in today's heavily researched approaches to NLP that don't lead to NLU. Understanding has now been well researched and models like Patom theory promise to exploit the knowledge we already have explaining how languages work.

REFERENCES

Ball, John. Patom Theory, 2017.
http://www.isi.hhu.de/fileadmin/redaktion/Oeffentliche_Medien/Fakultaeten/Philosophische_Fakultaet/Sprache_und_Information/Ball_PatomTheory.pdf

Ball, John, Speaking Artificial Intelligence, 2015.
http://www.computerworld.com/author/John-Ball/

Ball, John. Machine Intelligence: The Death of Artificial Intelligence, *Hired Pen Publishing*, 1998/2016.

Bar-Hillel, Yehoshua. The Present Status of Automatic Translation of Languages. *Advances in Computers, vol. 1*, p.91-163, 1960.
http://www.mt-archive.info/Bar-Hillel-1960.pdf

Chomsky, Noam. Syntactic Structures. *Mouton Publishers*, 1957.

Cortés-Rodríguez, Francisco José. Towards the Computational Implementation of Role and Reference Grammar: Rules for the Syntactic Parsing of RRG Phrasal Constituents, 2016.
http://pendientedemigracion.ucm.es/info/circulo/no65/cortes.pdf.

Firth, J.R. Studies in Linguistic Analysis. *Basil Blackwell*, 1962.

Gardner, Howard. The Mind's New Science: A History of the Cognitive Revolution. *Basic Books, Inc., Publishers*, 1985.

Guest, E. Parsing Using the Role and Reference Grammar Paradigm, 2009.
http://eprints.leedsbeckett.ac.uk/778/6/Parsing%20Using%20the%20Role%20and%20Reference%20Grammar%20Paradigm.pdf

Henaff, Mikael, Weston, Jason, Szlam, Arthur, Bordes, Antoine and LeCun, Yann. Tracking the World State with Recurrent Entity Networks. CoRR, abs/1612.03969, 2017.
http://arxiv.org/abs/1502.05698.

Jackendoff, Ray. Foundations of Language: Brain, Meaning, Grammar, Evolution. *Oxford University Press*, 2002.

McCord, Michael C. IBM Research Report: Using Slot Grammar. *IBM Research Division*, 2010.

Mikolov, Tomas et al. Distributed Representations of Words and Phrases and their Compositionality, 2013. https://papers.nips.cc/paper/5021-distributed-representations-of-words-and-phrases-and-their-compositionality

Miller, George A. WordNet: A Lexical Database for English. *Communications of the ACM Vol. 38, No. 11: 39-41*, 1995.

Pavey, Emma L. The Structure of Language: An Introduction to Grammatical Analysis. *Cambridge University Press*, 2010.

Pinker, Steven. The Language Instinct: How the Mind Creates Language. *Harper Perennial Modern Classics*, 1994.

Pustejovsky, James and Stubbs, Amber. Natural Language Annotation for Machine Learning. *O'Reilly*, 2013.

Pustejovsky, James and Bogurev, Branimir. Lexical Semantics: The Problem of Polysemy. *Clarendon Press*, 1996.

Santorini, Beatrice. Part-of-Speech Tagging Guidelines for the Penn Treebank Project, 1990. http://repository.upenn.edu/cgi/viewcontent.cgi?article=1603&context=cis_reports.

Seung, Sebastian. Connectome: How the Brain's Wiring Makes Us Who We Are. *Mariner Books Houghton Mifflin Harcourt*, 2012.

Stillings, Neil A. et al. Cognitive Science: An Introduction. *The MIT Press*, 1987.

Weston, Jason, Bordes, Antoine, Chopra, Sumit, and Mikolov, Tomas. Towards ai-complete question answering: A set of prerequisite toy tasks. CoRR, abs/1502.05698, 2015. http://arxiv.org/abs/1612.03969.

Van Valin, Jr., Robert D. Exploring the Syntax-Semantics Interface. *Cambridge University Press*, 2005.

Van Valin, Jr., Robert D. Exploring the Syntax-Semantics-Pragmatics Interface. *John Benjamins Publishing Company*, 2008, 161-178.

Van Valin, Jr., Robert D and Lapolla, Randy J. Syntax: Structure, Meaning and Function. *Cambridge University Press*, 1997.

## Appendix A. SCIENCE & ENGINEERING SUMMARY

The historical problem of the scalability of NLP systems led to the parallel development of rules-based systems and statistical/neural network systems. The details of these approaches and the analysis of why a meaning-based combinatorial system is superior to both is covered in the appendices below and our other related documents[12].

Our model incorporates the brain-based Patom Theory[13] and the linguistic framework, Role and Reference Grammar (RRG). The core of the system is a meaning-based network—a bit like George Miller's WordNet[14]. The matching function serializes/deserializes text to meaning by:

> converting serial language in the form of sequential word forms (each, in turn, comprising symbol sequences) into a packet of hierarchical sets of meaning; and the reverse.

Using layered sets of meaning as the representation is language-independent by definition, with generation from meaning to language controlled by the valid phrase constraints in the target language (NLG) and recognition controlled by the source language's words and phrases (NLU).

Pattern matching is not the trivial "here are five symbols in sequence", but rather a match of a list of sets (i.e. Patom Theory). Each set element can only be stored once and is therefore an atomic pattern (i.e. no smaller constituent—necessary for automated decomposition and learning).

- RRG, based on the study of languages (linguistics), connects three areas via its **linking algorithm**:
    - syntactic representation
    - semantic representation
    - discourse pragmatics
- Patom Theory, based on the cognitive sciences (especially theoretical neuroscience and computer science), enables the implementation of RRG with modern computer software techniques. The key differentiators to other methods include:
    - no parsing, no trees
    - no parts of speech (e.g. noun, verb, adjective), but instead uses semantic universals (referent, predicate, modifier)
    - no grammatical rules (e.g. transitive), but instead predicate characteristics (e.g. number of roles, polarity, verb class, semantic relations)
    - uses lists of patterns to efficiently match phrases (that are also lists of patterns)
    - disambiguates meaning based on predicate's associations

**And no combinatorial explosion**

Google's Parsey McParseface announcement[15] admits that even today's best parsing technology must deal with the artificially created combinations where: "It is not uncommon for moderate length sentences—say 20 or 30 words in length—to have hundreds, thousands, or even tens of thousands of possible syntactic structures. A natural language parser must somehow search through all of these alternatives, and <u>find the most plausible structure given the context</u>" (our underline).

Instead of finding "plausible structures", our method finds the validity based on meaning.

---

[12] Ball, "Linguistic Analysis", patent number: 8600736 and "Set-based Parsing for Computer-Implemented Linguistic Analysis", Publication number: 20170031893

[13] Patom theory postulates that brains store, match and use hierarchical, bidirectional linksets only. Linksets are linked patterns of sets and lists only, where lists are derived from sets. It aligns with observed neuron function.

[14] http://www.computerworld.com/article/2935578/emerging-technology/miller-s-wordnet-paves-the-way-for-a-i.html

[15] https://research.googleblog.com/2016/05/announcing-syntaxnet-worlds-most.html

## Appendix B.   REFOCUSSING NLP FOR AI: STOP PARSING

At the core of NLP today is the desire to parse sentences into grammatical trees using parts of speech. In 2016, Google announced a global tool to help parse—SyntaxNet and its English-specific version, Parsey McParseface[16] and now other languages[17], too. Worse, the claim was that this tool would help with the problem of NLU, but understanding language requires not only the syntactic structure to be correctly recognized, but the meaning of the words, as well.

Google isn't the only company focused on parsing. The IBM Slot Grammar (McCord, 2010) supports 15 parts of speech and a largish number of features such as person, gerund, feminine, time, and dative. The Slot Grammar currently supports 6 languages.

NLU requires human-like error correction, such as handling misspoken words, backtracking and using emphasis to change meanings. It requires the ability to make sense of new words and phrases dynamically, and therefore learn on the fly—all features that differ from the traditional parser role.

Computational complexity theory experts even argue that parsing can never be accomplished as it is NP-complete[18]—NP (for nondeterministic polynomial time) and therefore some people today consider lack of parsing to be the limiting step for AI-complete solutions because the combinatorial explosion is so large as to exceed the capability of processing engines.

Why put all this effort for parsing into a task that, even after 60 years of continuous effort, is only part of the solution for NLU and has never had human-like accuracy? Attempts at rules-based parsing have failed using every conceivable approach, corpus linguistics performed better, but was still found wanting, and now this solution using artificial neural networks has known limitations.

As part of the goal of NLP is NLU, why not aim at NLU in the first place? Google goes on to point out:

> "Humans do a remarkable job of dealing with ambiguity, almost to the point where the problem is unnoticeable; the challenge is for computers to do the same. Multiple ambiguities such as these in longer sentences conspire to give a combinatorial explosion in the number of possible structures for a sentence. Usually the vast majority of these structures are wildly implausible, but are nevertheless possible and must be somehow discarded by a parser."

We think there is a better way—don't parse, don't use parts of speech, and don't use trees. In order to understand: use the meaning of words and their sets of associated meaning elements.

We believe that the desire to parse comes from the Chomskyan revolution starting in the late 1950s. In this section, we refute the claim that syntactic structure can be investigated independently from meaning. Secondly, we refute the idea that parts of speech have a useful role in sentence analysis. Lastly, we suggest that trees aren't as effective as sets at representing sentences because sets simplify and reduce the number of phrase patterns needed.

When parsing a natural language using only rules and not meaning, an unsolvable combinatorial explosion is created. The alternative is our NLU model.

SYNTACTIC STRUCTURE WITHOUT MEANING
This is the argument for studying grammar independently of meaning by Noam Chomsky, "the

---

[16] https://research.googleblog.com/2016/05/announcing-syntaxnet-worlds-most.html
[17] https://algorithmia.com/algorithms/deeplearning/Parsey
[18] https://linguistics.stackexchange.com/questions/3629/can-parsing-be-classified-to-some-complexity-class-e-g-np-complete
for a discussion and some references.

father of modern linguistics" and one of the "most cited scholars in history[19]."

"The fundamental aim in the linguistic analysis of a language L is to separate the *grammatical* sequences which are the sentences of L from the *ungrammatical* sequences which are not sentences of L and to study the structure of the grammatical sequences" (Chomsky, 1957, p13)

This was the revolutionary aim of linguistics, back in 1957, the year after AI and cognitive science began, still having a profound effect today—even in companies focussed on artificial neural networks for AI, of which NLP is but one solution. The question is, what is a grammatical sentence? It can't be defined in terms of what a native speaker of the language would think, because those people have learned the language. Their brains have done what is needed to relate their recognition of objects and the manipulation of them to words and phrases, already, so they are effectively NLU systems. Wouldn't it be better to study how we use word sequences to convey meaning in context?

Let's continue the line of argument. "Customarily, linguistic description on the syntactic level is formulated in terms of constituent analysis (parsing)" (Chomsky, 1957, 26)

Interestingly, similar rules are needed to differentiate subtle differences, even in this early work, such as plural noun phrases having different rules to singular ones. Then it follows that different sentence phrases are needed to (a) combine plural subjects to plural verbs and (b) to combine singular subjects to singular verbs. Then more rules for first person singular subjects, to first person singular verbs, if needed. The more subtleties that are found in the language the more rules are needed—causing an explosion in rules.

But what is the connection between phrase structure, its rules and natural language independently to meaning?

> "The notion 'grammatical' cannot be identified with 'meaningful' or 'significant' in any semantic sense. Sentences (1) and (2) are equally nonsensical, but any speaker of English will recognize that only the former is grammatical.
> 1) Colorless green ideas sleep furiously
> 2) Furiously sleep ideas green colorless" (Chomsky, 1957, p15)

It is odd to look at a fully working NLU system (a "speaker of English" by definition) and then claim that the recognition of the patterns is in some way evidence that "meaning" is excluded. True, the semantics of the words don't make sense with the cherry-picked sentences, but the words have meaning in addition to pure semantics. Ideas is plural—meaning there is more than one. Sleep is a present tense, third person, singular form (it's obviously a verb). There are obvious phrases and a clause in this "grammatical" sentence. "What slept?" A: "Ideas". "How did they sleep?" A: "Furiously" Do our answers come from meaning, or just something separate, called grammar?

As our brain is the only known, accurate parser today (machine can't do it accurately) we can see how an English speaker breaks (1) up as:

JJ JJ NNS VBP RB[20] (adjective, adjective, plural noun, verb non-3rd person singular present, adverb)

It then groups them as JJ JJ NNS (type of ideas), then VBP RB (a way of sleeping) and then that the ideas sleep (NNS VBP). But how does my brain do this? Is my brain throwing away some word senses for different parses based on statistics? Based on consistency for the entire sentence?

---

[19] https://en.wikipedia.org/wiki/Noam_Chomsky
[20] Using the Penn Treebank POS tags (Santorini, 1990, 6-7)

LEARNING WORDS ACQUIRES MEANING

In the first place, we **learn** the meanings of words. If we were to completely swap the meanings of the words to align (2) with (1) with colorless=furiously, green=sleep, sleep=green and furiously=colorless. (2) becomes "grammatical" and (1) becomes ungrammatical. In computer software, the mapping from the word form to the representation must be specified. Surely the word learning process will connect the qualities (colorless, green) to representations in the brain that differ to the representation of ideas, that differ to the representation of sleep that differs to the representation of fury? Wouldn't the singular and plural forms of an idea be related to 'Number', a referent phrase operator to differentiate between singular and plural?

The point is, there is something about the words themselves that allows a native speaker to detect grammaticality. We claim that is meaning. Words convey meaning in sentences and you can't split it out. Some languages use different packaging; for example, French adds phonemes to verbs to indicate future tense, while English uses a specific word in a constrained sequence.

But let's assume that the grammaticality detected by a speaker of English is due to, say, a modern part-of-speech tag sequence being matched.

- Colorless green ideas sleep furiously
- **JJ** **JJ** **NNS** **VBP** **RB**

There is our starting point. Let's use our knowledge of grammars to create three new rules to finish our brains' grammatical parse. We will use Backus-Naur form[21] for the definitions.

NP ::= JJ JJ NNS (a noun phrase is two adjectives followed by a plural noun)
VP ::= VBP RB (a verb phrase is a present tense, plural, 3rd person verb plus an adverb)
S::= NP VP (a sentence is a noun phrase followed by a verb phrase)

So here, we recognize a sentence S, given the input that corresponds with the grammatical result shown. Now because this sentence is grammatical, and is independent of meaning, all English speakers will recognize any substitutions with equivalent grammar, because it is the grammar that matters, not the meaning. Let's carefully choose meanings we know are of the same type.

The running men = DT JJ NNS (determiner, adjective, plural noun). 'Running' is an adjective (JJ).

The running of the men = DT NN IN DT NNS (determiner, singular noun, preposition, determiner plural noun). Running is a singular noun (NN). But we can make running plural, by adding an 's'. The runnings of the men = DT NNS IN DT NNS (determiner, singular noun, preposition, determiner plural noun). Runnings is a plural noun (NNS).

The men run = DT NNS VBP (determiner, plural noun, present non-3sg verb). Run is a present tense, singular, non-3rd person verb.

So let's map our new words into the grammatical sentence:

- **JJ** **JJ** **NNS** **VBP** **RB**
- Running running runnings run furiously

We challenge a speaker of English to claim that "running running runnings run furiously" is a grammatical sentence!

Now, we excluded meaning as the basis for grammaticality, and relied on the Penn Treebank part-of-speech guidelines for our analysis. This result, however, violates our view that grammaticality is independent to meaning based on an English speaker because while this is a grammatical sequence, it is also clearly an ungrammatical sequence.

---

[21] https://en.wikipedia.org/wiki/Backus%E2%80%93Naur_form

This illustrates one of the reasons so many intermediate structures are created by parsers. It is simply not possible to study a language independently to meaning, because part of what creates grammaticality is the meaning our brain associates. Those JJs, NNSs and VBPs concepts just aren't in our brains alone. We believe that sets of meaning-based elements are sufficient and appropriate to model a language and our experience shows it to be effective at NLU tasks.

What of the alternative case, when we understand a foreigner talking without our grammatical model? "Me city, por favor. Train" This seems to mean something like where's a train to the city please? Why does meaning work, somewhat, without grammar?

Aren't language learning capabilities capacities we continue to apply? For example, the letters NE1410S would presumably need to be included in corpora today for recognition in NLP, but the decoding of the letters is phonetic with odd groupings: NE = any, 1=one, 4=for, 10=ten, S = ess. That is: "Anyone for tennis?" Surely language, and parsers don't need to cater to this underlying pattern and all like it in order to speak a language!

Given a very large set of meanings, the sample "grammatical" phrase (JJ JJ NNS VBP RB) will produce meaningful sentences. A larger set will be recognized as "grammatical" but clearly meaningless. And the biggest set of all will produce the majority of the workload for parsers—implausible, impossible or invalid out-of-context word sequences. Our work in NLU eliminates those invalid matches early, based on the meaning of the other words in a phrase, so the recognition step returns the correct meaning or meanings efficiently.

To recap, Chomsky's revolutionary 1957 linguistics approach claimed that there are two different systems running in a brain in parallel: (a) a grammatical system that recognizes the syntax used; and (b) a semantic system that recognizes meaning. Only syntax was deemed important for linguists, without a need to understand meaning because: "...we are forced to conclude that grammar is autonomous and independent of meaning, and that probabilistic models give no particular insight into some of the basic problems of syntactic structure" (Chomsky, 1957, 17).

We have just shown that grammar, when studied independently from meaning, produces both grammatical statements, and ungrammatical statements. Our conclusion is that instead of focussing on language without meaning, focus on a broader linguistics model. We choose RRG for this reason as it combines syntax, semantics and discourse-pragmatics.

PARTS OF SPEECH VERSUS SEMANTIC REPRESENTATIONS
Part of the reason for parsing's failures is the adoption of the parts-of-speech model, based on the pursuit of grammar. Word classes, that define noun and verb, are problematic. 'Destruction' is a noun (meaning destroy), but it has its own definition in a dictionary, separately to destroy:
- destroy, destruct: verb 1, "do away with, cause the destruction or undoing of"
- destruction, demolition, wipeout: noun 2, "an event (or the result of an event) that completely destroys something"

Both of these definitions (destroy/destruction) can be brought together for use on a machine. They are word forms for a predicate and clearly have the same meaning in the right context (the destruction of the bridge was bad, the bridge was destroyed badly).

Based on work across multiple languages, while there are no universal categories of parts of speech, there are universals for referents and predicates (Van Valin, 2008, 162-178).

Let's look at another example to stress this point. We showed how the word 'running' is an adjective, a noun and a type of verb. Each of these definitions has a common semantic meaning that could be represented with the predicate run', like: (Say 'someone is running', 'the running someone' and 'the running by someone')

> do' (someone, [run'(someone)])

So three different annotations in corpus linguistics are for one meaning/predicate. While parsers assign illogical combinations of predicates in noun positions and illogical ones for referents in predicate positions, we have consolidated the meanings of referents or predicates with appropriate operators and attributes. 'Running' is only a predicate, with multiple forms.

PATTERNS INSTEAD OF GRAMMAR
Our original language project used pattern matching to create a 1980s-style parser without dealing with meaning. To match and generate phrases, though, required the maintenance of a large number of phrases. At one stage, to deal with tenses, aspects, grammatical objects, polarity and a verb, there were 1,000 or so patterns. It worked for both recognition and generation, but was limited to just that particular sequence. There would need to be an increase in the number of phrases to add how expressions (a subset of adverbs/adverbials) perhaps doubling that number, or worse.

The decomposition of language (Jackendoff, 2002, 38-67), like the brain's distributed functions is well appreciated. Leveraging combinatorial principles goes hand-in-hand with decomposition. Today, we use around five patterns to recognize and generate any English verb form correctly dealing with tense, polarity, aspect and passivity. And in English, five word forms suffice for the verbs themselves[22]. As is to be expected with software development, we do more with less, over time .

SETS INSTEAD OF TREES
A constituent parse tree uses a set of rules to convert from text to a tree. As rules are matched, the constituents are replaced by the defined token. This has proven to be unsuccessful to generate accurate representations for any natural language even after 60 years of trying.

In the linking model we use, the sentence "The red train was ridden on by the man who likes me" has two embeddings. "the red train" (historically called a noun phrase) and "the man who likes me" (also a noun phrase). In our system, both of these are recognized, converted to their semantic representations, disambiguated, and their context details passed to the context engine to be stored ahead of the main phrase: "the train"(Undergoer)—ridden on (Predicate)—"the man"(Actor).

As a referent phrase is recognized, it is validated semantically. But a parser cannot isolate these three constituents when they are meaningless because there is no other source of truth other than the tokens, or statistical occurrences. Language doesn't work that way, but allows us to isolate valid meanings in sentences as a part of the whole.

Notice that while there may be other matched patterns, unless the entire pattern is matched, there is only a set of intermediaries. Language is a precise tool in which a number of checks and balances validate it.

At the point when the set of details—actor, undergoer and predicate—are passed over, there are other set considerations available to a meaning-based system that seem to have no solution in the statistical worldview.

To make that statement into a question, once uttered, a tag question is easiest by adding the tag, "wasn't it" next. This is constructed with the negation of the auxiliary verb plus a repetition of the subject in pronoun form. So "The red train was ridden on by the man who likes me, wasn't it?" is a question. Clearly the set of information recognized so far would be most useful if it retained the type of auxiliary present, the polarity, and the gender of the subject. Perhaps other elements are necessary. That is one of the lessons have have learned through experience.

Our model collects elements that have been matched into sets. Instead of rules, it relies on phrase patterns to be matched. When a phrase pattern is matched, one of the matched elements is usually retained. All elements are labelled one or more times. Following a match, the phrase controls the creation of its matched representation—an overphrase, if you will. An overphrase contains the labelled elements, retained attributes and operators. A phrase can allocate attributes or operators to control to future flow of matches. Effectively, the match of a phrase results in a set of labelled

---

[22] The choice of use of verb or predicate would take too long to address precisely. We know generally what the words mean, and to differentiate between predicate-verb, predicate-noun etc. is unnecessary for the moment.

elements.

There are two functions of phrases in the system: consolidation and predication. <u>Consolidation phrases</u> combine elements with labels and extend its attribute/operator set. A consolidation set performs a syntactic function by combining elements together, but they are not designed to combine more than a single predicate's constituents. A <u>predication phrase</u> converts a consolidation set into a set of disambiguated elements containing its semantic representation.

The RRG completeness constraint applies to each predicate, not just the sentence: "All of the arguments explicitly specified in the semantic representation of a sentence must be realized syntactically in the sentence, and all of the referring expressions in the syntactic representation of a sentence must be linked to an argument position in a logical structure in the semantic representation of a sentence." (Van Valin, 2005, 129-130).

For example, to match the phrase 'the old cat', a consolidation phrase matches the two words, 'the old', retains 'the' and labels 'old' as 'adjective'. This consolidated phrase matches with 'cat', retains 'cat', and migrates the labeled element 'old'. This phrase also adds the operator 'definite'. We add other attributes to the new set, like breadcrumbs in the forest, to ensure the normal match of such a referent phrase is not matched more than once, such as by 'the the the old cat'.

Now a predication phase matches the pattern 'referent'+'definite' (the match of the consolidated set 'the old cat'). The completeness constraint is now verified, and if it fails, the phrase is deemed not to be matched. At this point, it accesses the word-senses for the referent 'cat' and validates that all connected cases of 'old' are existing predicate associations. This is a WSD step that reflects the original predicating use that created the association. Where there is no direct match, other semantic associations are verified depending on the particular predicating association to leverage inheritance. Next the logical structure/semantic representation (LS) is generated and stored in this predicate set for later use in context. Where consolidated set elements contains LSs, they are carried forward so all context elements are added to context in the right order.

EFFICIENCY
To scale an NLP system, there are an infinite number of possible phrases to use with a large vocabulary. The lack of repetition of language is problematic: "Go into the Library of Congress and pick a sentence at random from any volume, and chances are you would fail to find an exact repetition no matter how long you continued to search… If a speaker is interrupted at a random point in a sentence, there are on average about ten different words that could be inserted at that point to continue the sentence in a grammatical and meaningful way." (Pinker, 1994, 77).

Our phrase matching system is very efficient. Only phrase patterns that can potentially match are checked. There are a few kinds of phrases—literals, consolidation and predicating. Literals match words and subsequent words. Movie titles, for example, are literal phrases such as the 'wizard of oz'. This allows the title to be matched in addition to the phrase about a type of wizard. A literal phrase is indexed off the word itself, currently the first word in the literal phrase. These are checked the first time only, as they can't be built up. Typically, excluding the embedded phrases that go through their own consolidation and prediction steps, consolidation phases are indexed against one of their common <u>attributes</u>. That is, when an attribute is within the set, its connected phrases will be tested. So with one of the predicating verb classes, its associated word-sense will contain that specific categorizing attribute. If that word-sense is in the consolidated set, it will attempt to match and a resulting disambiguated LS with verified constituents will be the result.

Comparing this approach to parsing, by only testing patterns that **can** match greatly reduces the effort. While rules based tokens were quite limited in count compared with the set of valid words, lots of rules needed to be tested in a single parse step after initial tokens are created. In an extreme case with thousands of different predicate types, our NLU will still test only phrases indexed against the current senses or their senses' attributes. The fewer senses that exist against a word, the fewer phrase patterns are to be tested - unlike a parser.

## Appendix C. COMBINATORIAL REDUCTION EXAMPLE: OPERATORS & ATTRIBUTES

There are a number of benefits in moving from a computational model to a pattern-matching one. Operators and attributes take sequential information and convert it into a set element. As sets are unordered, multiple sequences can be converted to set elements, with fewer permutations to consider. This section looks at the operator and attribute benefits only as it reduces combinations.

RRG postulates a constituent projection (syntax) and an operator projection (modifiers). RRG represents predicates, arguments and lexical modifiers differently from grammatical modifiers. If we were to parse instead of link, this would still vastly simplify the resulting parse trees. The resultant predicates and referents are made far easier to locate in the constituent projection, because the noise from operator and attribute words and inflections are in the operator projection. As the operator projection contains the operators, it is also easy to recognize the meanings contained at the relevant level.

### Tense/aspect/negation/voice

Tense/aspect/negation/voice are non-trivial patterns in that they aren't simply words, but sequences of words/inflections. For example, the passive voice can be signalled by the pattern [meaning(be-auxiliary)] + [predicate+(past participle)]. Examples that match are: "is eaten", "was chased" and, when set-matching, "was not/only/.. attacked" (in set matching, the consolidation set labels and embeds matched elements as constituents and retains an active element. Here 'was' or 'p:be'[23] is active).

RRG lists the possible operators (Van Valin, 2005, 9) in clauses and in nouns (Pavey, 2010, 188) and these are bidirectional—a pattern converts word sequences to operators, and operators convert to word or inflections in a target language.

A multi-lingual quality can be contrasted in English and French. The conjugation method for French can be found on language learning websites like this[24]; *How to Conjugate the Simple Future Tense in French* using the verb *parler* (to speak): Use the entire verb as the stem, adding "*-ai, -as, -a, -ons, -ez, -ont*" at the end.
*Je parler**ai**—Tu parler**as**—Il/Elle parler**a**—Nous parler**ons**—Vous parler**ez**—Ils/Elles parler**ont***

And in English:
*I **will** speak—You **will** speak—He/she **will** speak—We **will** speak—You **will** speak—They **will** speak*

So, in English, the word 'will' identifies the operator followed by an infinitive verb form, but in French the pattern is solely word-based; it takes the infinitive form of the word plus a suffix that agrees with the psa or privileged syntactic argument (Van Valin, 2005, 94-100) of the sentence that carries agreement with the verb in some languages. As generation of language is obviously language dependent (both word sequence and word selection are language-specific properties), what's the semantic representation needed to generate this conjugation?

For the referent (the pronoun), we need the major category, 'referent', and the relevant operators and attributes: e.g. 'pronoun', 'first person' and 'singular' and for the predicate we need the meaning (p:speak), and its operators/attributes: e.g. 'predicate', 'first person', 'singular' and 'future'.

Therefore the phrase looks like:

['referent', 'pronoun', 'first person', 'singular']+['predicate', 'first person', 'singular', 'future'] where the predicate phrase includes tense.

We can generate "I will speak" or "Je parlerai" for English or French, respectively, directly from the NLU's meaning based on the target language's specific patterns (NLG). Tense operates on the entire proposition, so in English the tense generator produces 'will speak' while for French it produces 'parlerai'.

---

[23] 'p:be' is the index to the semantic element that is the predicate including the English verb 'to be'.
[24] http://www.fluentu.com/blog/french/french-future-tense/

Now we can readily extend this approach to illustrate the English tenses/aspect/voice/negation for the verb 'speak'. An internet web site explaining the patterns for perfect, passive, progressive and negation will summarize the patterns needed.

So given 'future', 'passive', 'perfect', 'progressive', 'negative' with a meaning payload of 'p:speak', the generator outputs: "won't have been being spoken". Integrating this with the psa enables the additional operator, 'illocutionary force', to render the output as a question, command or statement, accordingly.

The purpose of table 9 (below) is to illustrate some of the range of word sequences that are built around three meanings (cat-rat-eat) with the extension of the RRG operator/attribute model. By reducing the words to predicates, referents and modifiers, we now have a triple that can extend the WordNet network, like a 'does-x' relation. Therefore, the specific association types in WordNet extend to include hypernym/hyponym/is-a, holonym/meronym/has-a and does-x).

### Table 9. Samples of generated text with operators/attributes

| Actor | Undergoer | Predicate | Output |
|---|---|---|---|
| **cat (tabby)**<br>definite | **rat (furry)**<br>indefinite | **eat (chew/swallow)**<br>past | The cat ate a rat |
| **cat (tabby)**<br>indefinite | **rat (furry)**<br>indefinite | **eat (chew/swallow)**<br>present | A cat eats a rat |
| **cat (tabby)**<br>indefinite | **rat (furry)**<br>indefinite | **eat (chew/swallow)**<br>present, negative | A cat doesn't eat a rat |
| **cat (tabby)**<br>indefinite | **rat (furry)**<br>indefinite | **eat (chew/swallow)**<br>present, statement, negative, passive, perfect, progressive | A rat hasn't been being eaten by a cat |
| **cat (tabby)**<br>definite, singular, deixis, proximal | **rat (furry)**<br>definite, plural, deixis, distal | **eat (chew/swallow)**<br>present, statement, negative, passive, perfect, progressive | Those rats haven't been being eaten by this cat |
| **cat (tabby)**<br>definite, singular, deixis, proximal | **rat (furry)**<br>definite, plural, deixis, distal | **eat (chew/swallow)**<br>future, question, negative, passive, perfect, progressive | Won't those rats have been being eaten by this cat? |
| **cat (tabby)**<br>indefinite | **rat (furry)**<br>indefinite | **eat (chew/swallow)**<br>present, negative, passive, question | Isn't a rat eaten by a cat? |

Finally, having removed a lot of 'noisy' elements from text (operators and embeddings), we are left with elements that actually comprise the focus of dictionaries (a bit like RDF[25] except RDF uses the ambiguous syntactic concepts of subject and object). These language-independent associations provide a new way to recognize and generate valid phrases in a source and target language, as the predicating step validates the word senses. The association can be a learning step, assuming you know what the meaning is in the first place, that can be refined with additional cases. For now, our team just creates such associations when we require WSD based on the logical human definition.

---

[25] https://www.w3.org/TR/2004/REC-rdf-concepts-20040210/#section-formal-semantics

Appendix D.   WHAT DO WE MEAN BY MEANING?

A word's meaning is its relevant definition. By automating the selection of a word's definition, or its word-sense, we have made a start towards NLU: finding the meaning of the words in a sentence based on the meanings of the other words in the sentence. For our purpose, a dictionary definition provides a starting point for meaning since, given a definition, we can modify it to reflect our NLU requirements. Our dictionary won't be based on parts-of-speech, but instead on the universal semantic elements: referents, modifiers and predicates. During the conversion, a lot of duplicated definitions will be deleted.

For example, again, 'running' is a predicate with a verb definition "move fast by using one's feet, with one foot off the ground at any given time". It is also a noun, "the act of running; traveling on foot at a fast pace". We delete this one. It is also an adjective, "done with a run". We delete this one, too. So we are left with one definition that is a one-role activity predicate.

### Intelligent Dictionary and Universal Encyclopaedia

An intelligent dictionary takes a short sample sentence that includes the word whose definition you want, and determines the possible definitions for you. "The young girl ran" will give you the definition above for 'ran'. "The water ran" will give you a different definition for 'ran', synonymous to 'flow'. The definitions of the words come from the meanings of the other words. It is 'intelligent' because the dictionary decides the meaning, not you, the user.

A current dictionary definition starts with a part-of-speech such as (n., adj., v., conj., etc) and shows any irregular word forms. Referents tend to be identified by the definition's first few words if they identify a category. For example, an 'eatery' is "a restaurant…", an 'eater' is "a person who eats", and 'ebony' is "a heavy hard very dark wood…". In other words, the definition of a referent often starts by identifying its more general definition, its hypernym relation. Other types of referent, like locations and temporal types provide a hypernym that identifies the category. For example: 'here' is "this place" and 'soon' is after "a short interval of time". The overlaps between traditional parts of speech (noun, adverb, adjective, verb) can duplicate the the definitions and also twist the meanings to the particular part of speech defined. Predicates tend to specify their entailing actions, 'eat' is to "take into the mouth, chew, and swallow (food)". Note that the undergoer's hypernym is also specified in the predicate in this case, 'food'[26].

Of course a definition is not the full answer, because there is also meaning embedded in the inflections of a word (RRG calls some of these modifiers, 'operators'). The difference between 'the cat' and 'the cats'? While both have the 'definite' operator, 'cat' also has the operator, singular, while 'cats' has the operator, 'plural'. The difference between 'eats', 'ate' and 'eaten'? The first has operators 'third person', 'singular' and 'present', the second has 'past' and the third is an attribute, 'past-participle' (not an operator, but in a phrase it resolves operators or other attributes  e.g. 'is being eaten' resolves 'present', 'passive' and 'progressive'). Languages have different packaging mechanisms for this information, as mentioned earlier.

So when we talk about meaning, we are referring to the appropriate definitions, within a semantic representation/logical structure/LS (these terms are treated as synonyms in this paper).

### WordNet - a semantic network

Dictionaries such as WordNet were designed for human readers. Humans have NLU, so by providing the necessary definitions and associations (such as hypernyms, actor and undergoer categories and entails relations), a human can learn the new vocabulary. Usually an example will cement the meaning better than a definition for someone. The template sentence allows the brain to store the relevant associations that are used without an intellectual process.

"WordNet® is a large lexical database of English. Nouns, verbs, adjectives and adverbs are grouped into sets of cognitive synonyms (synsets), each expressing a distinct concept. Synsets are

---

[26] The late Professor George A. Miller explained observations of dictionaries as his inspiration for aspects of the WordNet project.

interlinked by means of conceptual-semantic and lexical relations[27]... The most frequently encoded relation among synsets is the super-subordinate (transitive) relation (also called hyperonymy, hyponymy)… Meronymy, the part-whole relation holds between synsets...Parts are inherited from their superordinates...Parts are not inherited "upward" as they may be characteristic only of specific kinds of things rather than the class as a whole"

Most "relations connect words from the same part of speech (POS). Thus, WordNet really consists of four sub-nets ...with few cross-POS pointers. Cross-POS relations include ... (those) that hold among semantically similar words sharing a stem with the same meaning: observe (verb), observant (adjective) observation, observatory (nouns)."

Put another way, WordNet connects by using well-known sets of semantic associations, is-a, has-a, etc for English words. We add a missing one, something like does-x to enable WSD and some others to align with the other requirements of predicates.

To provide a resource for English, the design needs to be extended to align with our requirements for a language. Word forms are added to link their related word-senses with relevant attributes and operators. Other languages will have differing requirements, of course, but in theory dictionary definitions for AI can be language independent with this model. WordNets have been extended to multiple languages[28], but they are separate systems and none are yet in a format for NLU exploitation.

### Word-sense Disambiguation (WSD) - finding valid meanings

Our goal is to disambiguate meanings (known as word-sense disambiguation or WSD) in conversation to the extent needed to understand the source. There need not be full disambiguation, but some level of predicate resolution within a sentence is important. Lack of WSD has held back NLU for decades.

Consider these sample sentences: "the wind ate the mountain", "the girl ate the mountain" and "the girl ate the sandwich". What does 'eat' mean, and why, in each case?

Our approach to a predicate is to define it based on the human definition. The actor in the first case is inanimate, but moving. The undergoer is a non-living, non-food entity. That predicate is assigned a meaning (word-sense) and intersection association (does-x) possibilities: (actor) inanimate-motion - (predicate) p:eat1 and (undergoer) non-living, non-food. So given the wind at the mountain, the actor matches this pattern (wind is inanimate motion, i.e. it has a hypernym relation to this effect) and the undergoer also matches. As this specific predicate, 'eat', matches its actor and undergoer positions, it is a valid word sense. This match allows us to remove the other predicates ('eat' has a number of potential senses) and also remove all meanings of the actor and undergoer that don't match. This is the concept of WSD that is semantically-based. Languages allow the embedding of predications in sentences, and so the resolution of word-senses as early as possible is helpful for comprehension. The longer the sentence and the more embeddings, the easier it is to identify the valid word-senses.

What about "the girl ate the sandwich"? Here, the actor is an animal, and the undergoer is food for the predicate of eat meaning to "chew and swallow". The girl is an animal, and a sandwich is food, so this meaning is validated.

In the last case, "the girl ate the mountain" we get no matches. There is no form of the word 'eat' that we have defined, so the sentence is left as meaningless.

This is a method to confirm the word-senses in a predicate based on its association category. It aligns with Patom theory, because the decision to keep a sense or not is deferred to a simple match of a pattern. Does this undergoer match this category? No? Then this is the wrong predicate. Yes? Then this predicate is valid.

---

[27] https://wordnet.princeton.edu/wordnet/
[28] http://globalwordnet.org/wordnets-in-the-world/

There doesn't need to be a human-form dictionary definition for a machine to make use of this network. However, for diagnostic purposes, a written definition is helpful. An alternative would be to generate a definition from the defined associations.

If an NLU system chooses the right definitions automatically, based on the other words in the sentence, it is understanding. This is arguably what a human brain is doing in conversation in any case: recognizing the meaning of words based on the other words' meaning. For a system to be doing NLU, it must at least know the correct definitions of the words it is using. It must also map their associations in a semantic representation. Systems that produce the right results, without understanding the meanings of the words, are not doing NLU. Operators and attributes don't need a definition because their role conveys modifier information used in context.

The above meaning is the output of our meaning matcher. But to converse, there is more to come because we discuss things and then refer to them in context. To date, discourse-pragmatics, the third pillar of RRG, is probably the most neglected aspect of NLU.

LEARNING FOR DISAMBIGUATION
Patom theory claims that experiences are associated automatically, and that in brains, the specific defines the general[29] (i.e. we learn through experience). When we learn a sentence, we are learning predicates initially. Of course we are learning the grammar—how to recognize a referent, and how to recognize a predicate, too—but that's a bit easier as we have real experiences to tie them to (the multisensory objects or actions we experience).

And all we get is a single case of the predicate that makes sense. As the elements of our system are sets and lists, a predicate can be thought of as a set of associated roles. "Mary hit the ball" associates an action, 'hit', with an actor, 'Mary' and an undergoer 'the ball'. If the elements weren't sets, there would be little more to do now, but as they are, let's assume that all elements are associated via the predicate roles—an actor link and an undergoer link (bidirectional, of course).

Some time later, we get another match with the same predicate, say, "John hit the table". This is where things get interesting. The intersection of the actor position find both 'Mary' and now 'John'. The common elements of these are that they are singular and people—or whatever! On the undergoer side, they are physical objects, and singular. Over time, the best intersected patterns will probably be that animals (actor) hit things (undergoer). When the predicate matches these roles, it is a shortlist of valid meanings of the predicate, and also selects a subset of the meanings of the roles. This is a reasonable way to acquire the arguments of predicates based on experience.

At the core of intersection for predicates, therefore, are the associations through experience of the roles. When the specifics don't match, that predicate is wrong and others can be tried. As there are a number of possible verb classes, each may have its own definition resulting in accurate NLU in which matched sentences result in disambiguated context.

WHAT IS CONTEXT?
Patom Theory models brains as pattern-matching machines—storing, matching and using hierarchical, bidirectional linkset patterns. To learn, brains automatically connect patterns based on contiguity. Therefore, all experience is "connected" at a point in time, albeit in a hierarchy: visual receptors in the eyes never connect to a pattern of multi-sensory experience. To avoid the explosion of information, patterns operate autonomously, once learned by experience. In our NLU, we use the inspiration that a neuron basically accepts patterns as input, and when matched, the neuron activates. For the linguistic model, the set is matched when its inputs are matched.

The patom model is conceptual: it models brains, not neurons. While modeling neurons as set matchers seem to provide a better explanation than, say, as a computer processor, patom theory explains a number of brain capabilities, such as the ability to recognize patterns with different modalities.

---

[29] https://www.computerworld.com/article/2928992/emerging-technology/a-i-is-too-hard-for-programmers.html

Computer scientists tend to view information as something to create, retrieve, update, and delete (CRUD) in line with storage persistence. Brains should be viewed differently, as create and retrieve only. There is no deletion and no update. Something added to context cannot be deleted or changed. Discussions of context use in the computer industry has inadvertently abused the concept of context. That is unfortunate. Human memory is contextual, in the true sense of the word context [30]. For example, in an explanation of brain science, consider the concept that someone is no longer married. Does that mean the facts around the wedding and associated memories should be **deleted** (Seung, 2013, P84-85)? This doesn't align with our knowledge of how context works - especially in neural networks like our brain. Here, the idea that forgetting means to delete the associations runs counter to our knowledge. We know, for example, Lady Diana died and Prince Charles remarried. But we retain the context that at one time in the past, they were married. Context connects elements together and we never forget (until our relevant brain's elements physically fail).

We learn that "1+1=2" when we are young, but then as computer science students, we learn that "1+1=10". Both statements are true, but without knowledge of the context, one of those statements looks wrong (the second sum is in base 2, of course).

The meaning matcher performs the de-serialization function and outputs a semantic representation: a layered set of related elements as explained by RRG. To date, we use this as an element of context. The sequences of these context elements tells a story, but at any time, its constituents can be directly accessed through a pattern.

MEANING SUMMARY
The end result of this network: the recognition of a word-sense is possible by accessing the word form. Once a specific word sense is found (predicate/referent) it can generate a definition based on its predicate's roles or its referent's hypernyms. A "poodle" is defined as a "breed of dog with curly hair". This can be generated from "poodle" is-a "dog" and "poodle" has-a "curly hair". Similarly, "eat" can be defined as the sub-actions (a list) of "take into the mouth, chew, and swallow". But the actor position is "animal" and the undergoer is "food". Any undergoer that is considered food, is validated.

The benefit in defining such simplistic associations means that accuracy in NLU depends of a series of tests—match? Or not match?

## Appendix E.   OTHER MODELS LIKE DISTRIBUTIONAL SEMANTICS

We have spoken with many researchers: (a) inactive ones who tried and were demoralized by rules-based expert systems and translation systems, some running off linguistic frameworks needing POS rules and others needing statistics and then (b) active ones using skills to apply supervised and unsupervised variations of black-box technology for Deep Learning, statistical sources or combinations.

Words are meaningless symbols and it is their connection to meaning that enables human communications. While languages are built on a layered model (Van Valin, 1997, 49) that leads to the observation that similar words together are often related, the opposite is not true. Similar words in different contexts may not mean the same thing—even in large distributions on average. Equally, the same words in a sentence may not mean the same thing in different documents, either. Take an extreme case of the pronoun 'it': "The dog bit the cat. Did the dog bite the cat? Yes, it bit it." Here, each 'it' is unambiguously referred to in the question's narrow focus—dog and cat, respectively. The problem of the infinite nature of language has serious consequences for statistical modeling.

Some engineers have shown anger when confronted with the fact that meaningless words, even on a large scale, do not contain the meaning we use in language. An executive from one of the world's largest IT companies on earth explained to us that they truly believed that brains learn language with distributional semantics, and not associations of sensory experience. This is the consequence of the gap between the current NLP industry and the field of linguistics whose approach is currently out-of-favor because of past, persistent failures.

---

[30] Computer scientists redefined the meaning of context as "the minimal set of data that must be saved to allow a task to be interrupted, and later continued from the same point". Brains aren't like that.

## Distributional Hypothesis

Deep Learning systems for NLP and other machine learning applications need a lot of processing power and a lot of data to be effective. Andrew Ng, AI luminary, wrote this analogy: "To build a rocket (deep learning system) you need a huge engine (deep learning) and a lot of fuel (data)[31]."

But many papers come back to the concept of context, quoted as: "You shall know a word by the company it keeps!" (Firth, 1962, 11). This doesn't seem to anticipate the idea of context as meaning "proximity in documents" but human-like context as discussed by Wittgenstein. It's the opposite theory. It may well be an inspiration, but we doubt many normal human beings think of language the way proposed by distributional semantics. Most bAbI tests leverage, at their core, tools like word2vec (Mikolov et al, 2013) that seek to exploit this extended hypothesis.

A common problem today is around corpora, for which the bAbI tasks act like story corpora. What happens when there are no annotations? The creation of such input documents is an expensive issue and often a showstopper for these kinds of supervised learning systems. The other common problem is how to deal with system failures in corpus linguistics. The modification of corpora, or changing the algorithms both lead to a reload of the system.

The bag-of-words processing paradigm, and corpus linguistics in general, are clearly not brain-like because they don't deal with context or acquire language. A bag-of-words model, for example, bundles in linguistic operators and therefore loses that associated meaning that is fundamental to language. The 'bag of words' approach is used extensively in many NLP systems, because it works some of the time and is better than other options. Obviously, any scaled system based on the 'bag of words' principle will fail for English, as many English words get their meaning in a predicate based on their order.

For example, the predicate "on the beach" represents a physical location on the surface of an area of sand next to water. 'On' is a predicate:

be-on' (the beach, ∅)

where ∅ is the symbol representing the unspecified thing (argument) that is "on the beach". The thing could be a person, "The person is on the beach", or another predicate, "**The girl is happy** on the beach."

The reordering of the predicate loses its value: "the on beach" and "the beach on" don't mean "on the beach". Similarly, "**The girl is happy** on the beach." loses its meaning when shuffled: "The beach the happy girl is on." even though an extended sentence could restore a similar meaning: "The beach the happy girl is on **is in Tahiti**". Shuffling words is bad for English meaning, so the 'bag of words', even when it makes an application produce responses, is fundamentally not an NLU system and can never be one.

## NLU

While we know of context as a set of events that take place concurrently in experience, it is not the same as the word forms written in a small place. Semantics is a **set** of associations, while grammar is a **list**. Until syntax is converted to semantics there is no meaning. The human meaning of context allows us to interact appropriately with others based on the situation. Some computer scientists have redefined context to relate to the word forms in an area of corpora. As word forms are highly ambiguous in meaning, the conclusion that a set of similar word forms in one area of corpora means the same as that set in another area is a nice hypothesis, but clearly wrong based on language.

Our approach solves a number of important blockers present in other technologies and in a number of markets, providing a real pathway for improvement. Language learners could benefit from accurate interaction with a machine, for practice. IoT applications could benefit in having natural interactions, which are not possible with current technologies.

---

[31] https://medium.com/nanonets/nanonets-how-to-use-deep-learning-when-you-have-limited-data-f68c0b512cab

WordNet, or another similar resource, has yet to be converted to a meaning-based resource because lack of diversity in AI has led to ignoring that approach in favor of an all-in model for the distributional hypothesis. Current technological advances (1980s' computer tools available to Miller bear little similarity to today's) mean that Professor Miller's experiment could have been scaled to provide a global service for natural languages, unlike the multiple WordNets built around the world as language-specific repositories without scalability due to their old design limitations. (Imagine building a lexical database today in which there is no direct access, no independent index, to the meaning without first finding another element!) Miller's grand experiment has led the way to new systems, but only once we focus resources on meaning-based systems.

In the end, the best ideas win out, and it is rare that one idea wins at all problems. It's like scientific progress as better epicycles won against heliocentric models for centuries, until it failed.

## Questions Needing Answers

This section has a reasonably high level set of questions that seem like they can never be solved by models like distributional semantics.

**Question 1:** The motivation for a lot of modern NLP systems are based on 1950s-1990s concepts that required a lot of processing power. A universal encyclopedia was proposed to solve problems that to some degree are now a part of the bAbI tests, but it was dismissed as not being feasible at the time. For example, in 1960 Bar-Hillel rejected the potential of fully automatic machine translation for this reason: "...it is very easy to show its futility. What such a suggestion amounts to, if taken seriously, is the requirement that a translation machine should not only be supplied with a dictionary but also with a universal encyclopedia. This is surely utterly chimerical and hardly deserves any further discussion" (Bar-Hillel, 1960, 42).

A lot has changed since the 1950s. The reaction to the failure of rules-based systems seems to have left no alternative other than statistical methods. Word2vec (and other word-embedding approaches like GloVe, etc.) comes from a "research area of distributional semantics"[32]. They map words or phrases from the vocabulary to vectors of real numbers. The meaning of a word is the statistics of a vast combination of words in corpora. This approach is seen to fit in well with a number of current initiatives relating to neural networks. But the meaning of a word is its known associations, and future ones that can be taken from new experiences.

A queen (royalty sense), for example, is a female and a person. A king (again, as royalty) is a male and a person. A doctor and a nurse are both people. But corpus-based analysis will bias these genders based on probability. It is a very different way to look at language.

**Question 2:** Will a statistically focussed distributional system ever handle the scope of meaning that we believe is in this sentence: "The old sleeping cat a dog had been chasing today will try to go to the beach with her master". We think that an NLU system should produce the following contextual, semantic elements (table 10).

### Table 10. NLU Analysis: Impossible with many current models

| Logical Structure/ Semantic Representation (add to context) | Operators/ Attributes | Input to analyze |
|---|---|---|
| Input received | | The old sleeping cat a dog had been chasing today will try to go to the beach with her master |

---

[32] https://en.wikipedia.org/wiki/Word_embedding

| | | |
|---|---|---|
| **be'**(cat, [**old'**]) | **definite** | The old sleeping cat = <br> The cat is old (now intersect the predicate old with the undergoer cat) |
| **do'**(cat, [**sleep'**(cat)]) | **definite progressive present** | The old sleeping cat = <br> The (same) cat is sleeping (now intersect the predicate sleep with the undergoer cat) |
| dog | **indefinite** | a dog |
| **today'**( <br> **do'**(dog,[**chase'**(dog,cat)])) <br> **Call this resolved LS ϒ0** | **indefinite perfect past progressive** | a dog had been chasing the cat today (now intersect the predicate chase with the actor dog and the previously intersected context element, the undergoer cat) |
| **have'**(cat, master) | | her master (resolve the pronoun 'her'—two choices—cat and dog). There is only one referent <u>at</u> <u>this</u> <u>level</u>, so accept 'cat'. Assumes the communication is not to deceive (now intersect the predicate have with the actor cat and the undergoer master) |
| **do'**(cat and master,[**go'**(cat and master)]) & INGR **be-at'**(beach, cat and master) <br> **Call this resolved LS ϒ1** | | The cat - go to the beach with her master (intersect with 'master' to determine whether this is a second actor, a second undergoer or an instrument. Here, master isn't an instrument of going, and the activity predicate has no undergoer, so master is a second actor) |
| **do'**(cat,,[**try'**(cat, ϒ1)]) | **modal future** | The cat will try to "go to ..." (juncture embeds LS ϒ1) |
| **do'**(**today'**(**do'**(dog,[**chase'**(dog,cat)])),,[**try'**(cat, **do'**(cat and master,[**go'**(cat and master)]) & INGR **be-at'**(beach, **today'**(**do'**(dog,[**chase'**(dog,cat)])) and master))]) | **modal future** | OK, now to spell it out, here is the full semantic representation (with relevant operators and attributes as shown above). Due to the terseness of the representation, we normally show it as nested elements and drill down as needed, dynamically. |

The expected results are shown 'as recognized', noting that embedded phrases are added to context ahead of the composing phrases, since they must be resolved first in order to make sense.

After looking at the semantic representation of the above sentence, there are many words in proximity that relate to context. Of course they relate to the context, because humans speak in context. But whether or not there is a particular event taking place, human languages permit any language to be used, so while a lot of the time, the company of a word relates to its context, that's not truly specific. If your life depended on a particular sentence to be about atomic power at, say, a parliament whose representatives were discussing the cost of atomic power, would you take the bet?

Our NLU system has implemented the features shown above today and while it is not yet complete for any language, the benefits appear most promising.

There is no such thing as limited context in the real world. At a political forum, members may discuss soccer, their divorce or the history of Sparta. They may discuss a French play or the Prime Minister's toupee.

Even in a tightly controlled event, say, someone talking about the prime minister's coat, or friends, or their political views is common as an interruption to atomic power funding. Now looking at the constraints on the words used in the sentence above, it is clear that such information as contained in the semantic representation will be useful communications tools that are more than just 'proximity-based'.

**Question 3:** Lexical Semantics (Pustejovsky, 1996) explains the use of semantic associations to resolve meaning when inference is needed. How can the cases recognized in Lexical Semantics be resolved?

In a sentence like "Mary started the car" the unnamed association is that the car <u>has</u> an engine and the engine started. Intersection of the predicate 'started' with the undergoer 'car' will fail, but if we assume that a referent is the sum of its parts, intersection can also check those constituents. (this relation is 'has-a'.). By activating, on a missed match, the has-a relationships and intersecting again, we get a match on engine-start. Obviously this won't be appropriate for all predicates and all referents but, when valid, simply associating a unique attribute with the predicate will enable the action.

Another example of this kind is, "I'm parking out back". Here 'I' doesn't park, but if I 'has-a' car, that association that the car is parked out the back requires no additional semantic design.

The observations are that associations of meaning allow sentences to omit what would otherwise be the main ideas, such as the conversion of an unspecified predicate "John began reading the book" with "John began the book" where 'begin' is not the action, but instead a qualia of book is—'reading' (or 'writing'!) (Pustejovsky and Boguraev, 1996, 10).

In the second sentence, to get the same meaning, first the intersection must fail on began-(undergoer)book, which it will semantically as this referent has no starting point. But with the bidirectional link created from predicate to undergoer, book will have associations to what acted upon it—such as reading (what it is for) and writing (what created it). This is a minor change to the linkset intersection model during predicate casting (the combination of roles in a predicate includes the resolution of validity of arguments with the semantic network).

**Question 4**: Consider this starting sentence: "The cat ate the rat". We can change the sentence easily: "The cat ate, no, I mean it chewed on, the rat". Or further with: "The cat the dog chased at the beach today because it was playful ate, no, I mean it chewed on, the rat". These sentences demonstrate that some of the words used are not even part of the communications. They are just "errors". But distributional systems package these words up anyway, because you can't understand when you are simply creating statistics. How do Deep Learning systems propose to deal with this type of text where the premise that words carry meaning based on proximity is violated?

The common theme underlying all of these sentences is that the phrases in the sentence must first be recognized in order for meaning to be extracted. It isn't parsing, but recognizing the predicate groupings based on meaning. Set-based linking allows these examples, like a change in wording, to be addressed on the fly. Systems based on the theory of distributional semantics fundamentally cannot.